\documentclass[10pt,journal,compsoc]{IEEEtran}
\pdfoutput=1 
\ifCLASSOPTIONcompsoc
\usepackage[nocompress]{cite}
\else
  \usepackage{cite}
\fi
\usepackage{algorithm}
\usepackage{algpseudocode}
\usepackage{epsfig}
\usepackage{graphicx}
\usepackage{amsfonts}
\usepackage{comment}
\usepackage{caption}
\usepackage{color}
\usepackage{amssymb}
\usepackage{amsmath}
\usepackage{amsthm,mathtools,mathrsfs}
\usepackage{multirow}
\usepackage{etoolbox}
\usepackage{balance}
\usepackage{booktabs}
\usepackage{enumitem}
\usepackage{pifont,booktabs}
\usepackage[table]{xcolor}

\usepackage[colorlinks = true,
    linkcolor = red,
    urlcolor  = black,
    citecolor = blue, 
    anchorcolor = magenta]{hyperref}

\usepackage{tikz}

\usepackage{caption}
\usepackage{boldline}
\usepackage{makecell}%To keep spacing of text in tables
\setcellgapes{1pt}%parameter for the spacing
\newcommand{\cmark}{\ding{51}}
\usepackage{lineno}
\usepackage{adjustbox}
\usepackage{ dsfont }

\AfterEndEnvironment{lemma}{\noindent\ignorespaces}

\newtheorem{definition}{Definition}[section]
\AfterEndEnvironment{definition}{\noindent\ignorespaces}

\AfterEndEnvironment{example}{\noindent\ignorespaces}

\ifCLASSINFOpdf
\else
\fi
\begin{document}

\title{ARRQP: Anomaly Resilient Real-time QoS Prediction Framework with Graph Convolution}

\author{Suraj Kumar, 
    Soumi Chattopadhyay,~\IEEEmembership{Member,~IEEE}
\thanks{
Suraj Kumar and Soumi Chattopadhyay are with the Dept. of CSE, Indian Institute of Technology Indore, Madhya Pradesh 453552, India. 
(email: {\{phd2301101002, soumi\}@iiti.ac.in}).\\
Corresponding author: Soumi Chattopadhyay.\\
This work has been submitted to the IEEE for possible publication. Copyright may be transferred without notice, after which this version may no longer be accessible.
}
}

% \markboth{IEEE Transactions on Services Computing}{
% Suraj Kumar and Soumi Chattopadhyay: Anomaly Resilient Real-time QoS Prediction Framework with Graph Convolution}

\makeatletter
\long\def\@IEEEtitleabstractindextextbox#1{\parbox{0.922\textwidth}{#1}}
\makeatother

\IEEEtitleabstractindextext{
\begin{abstract}
In the realm of modern service-oriented architecture, ensuring Quality of Service (QoS) is of paramount importance. The ability to predict QoS values in advance empowers users to make informed decisions, ensuring that the chosen service aligns with their expectations. This harmonizes seamlessly with the core objective of service recommendation, which is to adeptly steer users towards services tailored to their distinct requirements and preferences. 
However, achieving accurate and real-time QoS predictions in the presence of various issues and anomalies, including outliers, data sparsity, grey sheep instances, and cold start scenarios, remains a challenge. 
Current state-of-the-art methods often fall short when addressing these issues simultaneously, resulting in performance degradation.
In response, in this paper, we introduce an anomaly-resilient real-time QoS prediction framework (called ARRQP). Our primary contributions encompass proposing an innovative approach to QoS prediction aimed at enhancing prediction accuracy, with a specific emphasis on improving resilience to anomalies in the data.
ARRQP  utilizes the power of graph convolution techniques, a powerful tool in graph-based machine learning, to capture intricate relationships and dependencies among users and services. By leveraging graph convolution, our framework enhances its ability to model and seize complex relationships within the data, even when the data is limited or sparse. ARRQP integrates both contextual information and collaborative insights, enabling a comprehensive understanding of user-service interactions. By utilizing robust loss functions, this approach effectively reduces the impact of outliers during the training of the predictive model. Additionally, we introduce a method for detecting grey sheep users or services that is resilient to sparsity. These grey sheep instances are subsequently treated separately for QoS prediction.
Furthermore, we address the cold start problem as a distinct challenge by emphasizing contextual features over collaborative features. This approach allows us to effectively handle situations where newly introduced users or services lack historical data. Experimental results on the publicly available benchmark WS-DREAM dataset demonstrate the framework's effectiveness in achieving accurate and timely QoS predictions, even in scenarios where anomalies abound.
\end{abstract}

}

\maketitle

\IEEEraisesectionheading{
\section{Introduction}\label{sec:intro}}
\noindent
In today's service-oriented digital landscape, ensuring Quality of Service (QoS) is a critical imperative. QoS prediction \cite{zibinSurvey2022}, the process of forecasting service performance, is indispensable for users and systems to make informed decisions. However, this task is riddled with challenges, including data sparsity, the cold start problem, the presence of outliers, and grey sheep instances.

Collaborative filtering (CF) \cite{ghafouriSurvey2022} has emerged as a highly promising solution for QoS prediction. However, early CF-based methods \cite{wsrec,nrcf,recf,racf}, which rely on exploring the similarity of users/services, suffer from low prediction accuracy due to their lack of consideration for data anomalies.

Despite some advancements using low-rank matrix factorization \cite{nmf,cmf,nimf,wramf,emf,csmf} and factorization machines \cite{afm,cadfm,efm} to tackle data sparsity, the cold start problem, and scalability, these methods frequently encounter challenges in achieving satisfactory performance. This is primarily because they have limited capacity to capture higher-order features and are incapable of handling various other anomalies such as grey sheep \cite{GS} and outliers \cite{cmf} present in QoS data.

Among recent advancements on QoS prediction, learning-based methods \cite{cnr,dafr,ncrl,cahphf,offdq,hsanet,trqp,tan,ldcf,cmf,clusterAE} are particularly notable for their performance improvement, as they excel in capturing the complex features of users and services.
Nevertheless, the learning-based methods that heavily rely on contextual features  \cite{dafr,dnmm,lafil,lbr,msdae} for QoS prediction may struggle to achieve desirable prediction accuracy when collaborative QoS features, which are often more relevant, are absent or not adequately considered. To further enhance prediction accuracy, recent techniques in the literature have aimed to address anomalies in QoS prediction, including handling outliers using robust outlier-resilient loss functions \cite{hampel1986robust,dclg,ldcf,cmf,hsanet} and addressing the cold start problem \cite{geomf}. However, many of these methods tend to focus on a subset of these challenges and may fall short when multiple issues coexist simultaneously, leading to performance degradation. Consequently, there is a pressing need for an innovative approach that not only enhances prediction accuracy but also bolsters resilience to anomalies in the data, which is the primary objective of this paper.

Our earlier work on QoS prediction \cite{trqp} introduced a graph-based solution to mitigate the issue of sparsity but did not address other challenges. The present research is an extension of our earlier work. Here, we not only enhance our predictive model but also prioritize the resolution of other anomalies, including outliers, the cold start problem, and the presence of grey sheep instances. Consequently, we have attained a substantial improvement in both prediction time and accuracy compared to our earlier work.

In this paper, we propose a scalable, real-time QoS prediction framework (ARRQP) that effectively addresses multiple anomalies, including outliers, data sparsity, grey sheep instances, and cold start, ultimately enhancing prediction accuracy.
Our proposed framework consists of five key components: two anomaly detection blocks responsible for identifying grey sheep users/services and outliers, and three prediction blocks. The first prediction block is engineered to withstand outliers and data sparsity for regular users and services. The other two prediction blocks are specifically dedicated to predicting QoS for grey sheep users/services and newly added users/services that lack sufficient data.
The primary contributions of this paper are outlined as follows:

\begin{itemize}[leftmargin=*]
    \item[(i)] We propose a multi-layer multi-head graph convolution matrix factorization (MhGCMF) model complemented by an outlier-resilient loss function. This model is designed to capture the intricate relationships among QoS data, effectively mitigating the influence of outliers. It not only enhances prediction accuracy but also ensures minimal prediction time, making it a valuable addition to QoS prediction methodologies.
    \item[(ii)] The synergy between contextual and QoS features, in addition to the spatial features automatically extracted by MhGCMF, significantly enhances the model's expressiveness in capturing the complex, higher-order association between user and services. This enhancement eventually leads to improved prediction performance.
    \item[(iii)] We introduce a sparsity-resilient method for detecting grey sheep users or services.
    \item[(iv)] Grey sheep instances possess unique characteristics, making it challenging to predict QoS values for such users or services using collaborative filtering alone. Consequently, we have devised a distinct QoS prediction model specifically tailored to address grey sheep users or services. This model incorporates a quantitative distinction measure obtained from the grey sheep detection block, allowing us to provide more accurate predictions for this particular category of users or services.
    \item[(v)] We address the cold start problem by designing a separate model that prioritizes contextual features over collaborative features.
    \item [(vi)] We conducted comprehensive experiments using the benchmark WS-DREAM RT and TP datasets \cite{WSDREAM} to assess the effectiveness of each block within ARRQP, as well as to evaluate the overall performance of ARRQP.
\end{itemize}

The rest of the paper is organized as follows. 
Section \ref{sec:preliminaries} presents an overview of the problem with its formulation. Section \ref{sec:method} then discusses the proposed solution framework in detail. The experimental results are analyzed in Section \ref{sec:result}, while the literature review is presented in Section \ref{sec:related}. Finally, Section \ref{sec:conclusion} concludes this paper.

\section{Overview and Problem Formulation}\label{sec:preliminaries}
\noindent
In this section, we discuss an overview of the QoS prediction problem followed by our problem formulation. We are given:

\begin{itemize}
 \item A set of $n$ users ${\mathcal{U}}$
 \item Contextual information $\mathcal{C}^u$ of each $u \in {\mathcal{U}}$, this contextual information includes user id, user region, autonomous system, etc.
 \item A set of $m$ web services ${\mathcal{S}}$
 \item Contextual information $\mathcal{C}^s$ of each $s \in {\mathcal{S}}$, where contextual information comprises service id, service region, service provider, etc.
 \item A QoS parameter $q$
 \item A QoS log matrix $\mathcal{Q}$ of dimension $n \times m$ containing past user-service interactions in terms of $q$, as defined:
 \begin{equation}
    \mathcal{Q} = 
    \begin{cases}
        q_{ij} \in \mathbb{R}_{>0} , & \text{value of $q$ of $s_j$ invoked by $u_i$} \\
        0 , & \text{otherwise}
    \end{cases}
\end{equation}
\end{itemize}

\noindent
% Table \ref{tab:qos_log_example} shows an example of the QoS log matrix of dimension $8 \times 9$, considering response time as the QoS parameter.
Each non-zero entry ($q_{ij} \ne 0$) in the matrix represents the value of $q$ of $s_j$ invoked by $u_i$. Zero entries, on the other hand, denote no interactions between user-service. It may be noted that the QoS log matrix is, in general, a sparse matrix.

% \begin{table}[!h]
%     \centering
%      \caption{Example QoS log matrix for response time (sec)}
%     \begin{tabular}{c|c|c|c|c|c|c|c|c|c} 
%     \hline
%               & {$s_1$} &  {$s_2$} &  {$s_3$} &  {$s_4$} & {$s_5$} & {$s_6$} & {$s_7$} & {$s_8$} & {$s_9$} \\ \hline\hline
%         $u_1$ & 0.0 &  0.0 &  0.5 &  0.0 & 0.0 & 2.5 & 0.0 & 0.0 & 2.2  \\ \hline
%         $u_2$ & 0.4 &  0.0 & 0.0 & 0.1 & 0.0 &  0.0 & 0.0 & 20.0 & 0.0  \\ \hline
%         $u_3$ & 4.0 &  0.0 &  0.3 &  0.0 &  0.0 & 0.5 &  0.0 &  0.8 & 0.0  \\ \hline
%         $u_4$ &0.0 &  11.5 &  0.0 & 0.0 &  0.9 &  0.3 & 0.0 &  0.0  & 3.9  \\ \hline
%         $u_5$ & 1.2  & 0.0 & 0.2 &  0.0 &  0.0 & 0.5 &  0.0 &  0.1 & 0.0   \\ \hline
%         $u_6$ & 0.0 &  5.0 &  0.0 &  4.5 &  0.0 &  3.0 &  0.0 &  0.2 & 7.0 \\ \hline
%         $u_7$ & 1.4 &  0.0 & 0.0 & 3.1 & 0.0 &  0.0 & 0.0 & 5.3 & 0.0  \\ \hline
%         $u_8$ & 0.0 &  0.0 & 2.6 & 0.1 & 0.0 &  0.0 & 0.0 & 1.2 & 0.0  \\ \hline
%     \end{tabular}
%     \label{tab:qos_log_example}
% \end{table}

The objective of the QoS prediction problem is to predict the QoS value of a given target user-service pair $u_i$ and $s_j$, where $q_{ij} = 0$. In general, the classical QoS prediction problem aims to reduce the prediction error as much as possible. However, minimizing the prediction error is challenging due to the following:

\begin{itemize}[leftmargin=*]
 \item {\emph{Data sparsity problem}} (${\mathds{S}}$): As we discussed earlier, the QoS log matrix is highly sparse. Therefore, minimizing the prediction error while predicting a missing value in the presence of other missing values is a challenging task.
 
 \item {\emph{Presence of outliers}} (${\mathds{O}}$): The presence of outliers in the QoS log matrix impedes minimizing the prediction error. Therefore, identifying and handling the outliers is essential to meet the objective.
 
 \item {\emph{Presence of grey sheep users/services}}  (${\mathds{GS}}$): The grey sheep users/services are the ones having unique QoS invocation patterns in terms of $q$. Therefore, predicting the QoS values of the grey sheep users/services as compared to the other users/services is difficult.
 
 \item {\emph{Cold-start problem}}  (${\mathds{C}}$): This is the situation when new users/services have been added to the system. Due to the absence of any past data, it is difficult to predict the QoS values of the new users/services.
\end{itemize}

\noindent
{\textbf{Objective}}:
This paper aims to design a real-time and scalable QoS prediction framework by addressing the above challenges to attain reasonably low prediction error.

\section{Methodology}\label{sec:method}
\noindent
In this section, we discuss our framework, ARRQP, for QoS prediction. ARRQP comprises two major anomaly detection blocks: (a) {\textbf G}rey sheep user/service {\textbf D}etection block (GD), (b) {\textbf O}utlier {\textbf D}etection block (OD); and three major prediction blocks: (a) {\textbf S}parsity and {\textbf O}utlier {\textbf R}esilient {\textbf R}eal-time {\textbf Q}oS {\textbf P}rediction block (SORRQP), (b) {\textbf G}rey sheep users/services {\textbf R}esilient {\textbf R}eal-time {\textbf Q}oS {\textbf P}rediction block (GRRQP), (c) {\textbf C}old-start {\textbf R}esilient {\textbf R}eal-time {\textbf Q}oS {\textbf P}rediction block (CRRQP). In the following subsections, we discuss each of these blocks in detail. We begin with discussing our first prediction block, SORRQP, which primarily deals with sparsity and outliers. 

\begin{figure*}[!h]
    \centering
    \includegraphics[width=\textwidth]{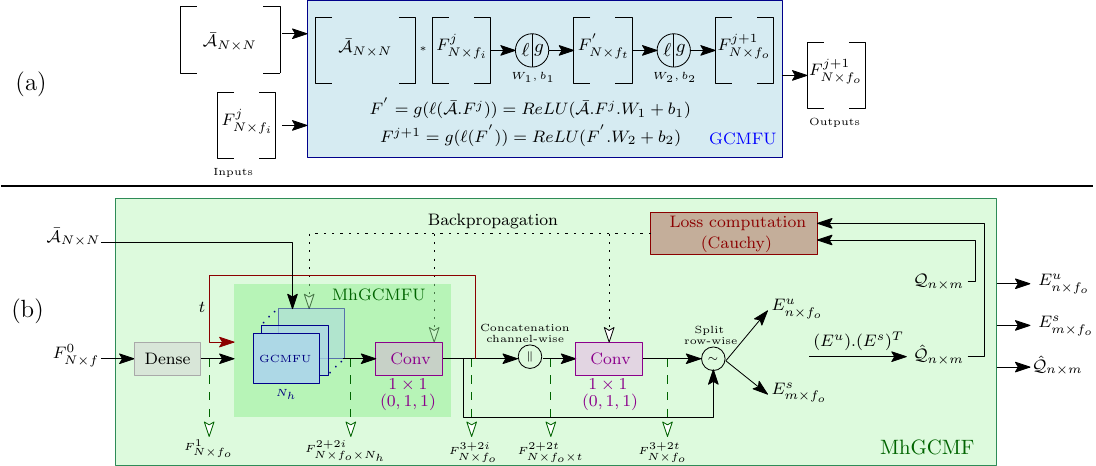}
    \caption{Architecture of SORRQP: (a) Details of GCMFU; (b) Architecture for MhGCMF}
    \label{fig:mhgcn}
\end{figure*}

\subsection{SORRQP Block} \label{sec:sorrqp}
\noindent
SORRQP leverages graph convolution \cite{gcn} to deal with the data sparsity. The graph convolutional network (GCN) takes privilege in graph architecture and effectively aggregates the node features through message passing between neighboring nodes. Graph architecture has more expressive power \cite{gcn} than most known representations of users and services. Therefore, in this paper, we adopt graph convolution as the fundamental operation in the SORRQP architecture. Here, we propose a multi-layer multi-head graph convolution matrix factorization (MhGCMF) model for QoS prediction. Fig. \ref{fig:mhgcn} shows the overview of SORRQP architecture. Before discussing the further details of MhGCMF, we first define the QoS Invocation Graph (QIG), which is a basic building block of graph convolution operation. 

\begin{definition}[QoS Invocation Graph (QIG)]
A QIG, $\mathcal{G}$ = ($V_1 \cup V_2$, $E$, ${\mathcal{E}}_1 \cup {\mathcal{E}}_2$) is a bipartite graph, where $V_1$ and $V_2$ are the set of vertices representing the set of users and services, respectively. An edge $e_{ij} = (v_i^1 \in V_1, v_j^2 \in V_2) \in E$ exists in $\mathcal{G}$ if $q_{ij} \ne 0$ in the QoS log matrix ${\mathcal{Q}}$. ${\mathcal{E}}_1$ and ${\mathcal{E}}_2$ represent the set of feature embedding for each node in $V_1$ and $V_2$, respectively.  \hfill$\blacksquare$
\end{definition}

\noindent
We now illustrate the details of the feature embedding used in this paper.

\subsubsection{Construction of the Feature Embedding} Our initial feature embedding comprises a set of QoS features along with a set of contextual features. We now briefly discuss the details of this embedding.

\noindent
{\textbf{QoS features:}} The QoS features include three different types of features, which are discussed below.
   
   \noindent
 {\emph{(i) Statistical features (${\mathcal{F}}^u_t, {\mathcal{F}}^s_t$):}} To capture the self characteristics of each user $u_i$ and service $s_j$, we compute 5 statistical features, as shown in Table \ref{tab:feature_vector}. It may be noted that since ${\mathcal{Q}}(i)$ is a partially filled QoS invocation vector (QIV), the data sparsity affects the statistical features.

\begin{table}[!h]
    \centering
    \caption{Feature embedding for QIG}
    \begin{tabular}{l|l}
    \hline 
        $min^u_{i} : min({\mathcal{Q}}(i))$ & $min^s_{j}: min({\mathcal{Q}}^T(j))$  \\ \hline
        $max^u_{i} : max({\mathcal{Q}}(i))$ & $max^s_{j}: max({\mathcal{Q}}^T(j))$ \\ \hline
       $\mu^u_{i} : mean({\mathcal{Q}}(i))$ & $\mu^s_{j}: mean({\mathcal{Q}}^T(j))$ \\ \hline
        $med^u_{i} : median({\mathcal{Q}}(i))$ & $med^s_{j}: median({\mathcal{Q}}^T(j))$  \\ \hline
        $\sigma^u_{i} : std\_dev({\mathcal{Q}}(i))$ & $\sigma^s_{j}: std\_dev({\mathcal{Q}}^T(j))$ \\ \hline   
%         \multicolumn{2}{c}{}\\
        \multicolumn{2}{r}{$std\_dev:$ standard deviation} \\
        \multicolumn{2}{r}{${\mathcal{Q}}(i)$: $i^{th}$ row of ${\mathcal{Q}}$; ${\mathcal{Q}}^T$: Transpose of ${\mathcal{Q}}$}\\
        
        % \multicolumn{2}{r}{}
    \end{tabular}
    \label{tab:feature_vector}
\end{table}

\noindent
    {\emph{(ii) Collaborative features (${\mathcal{F}}_n^u, {\mathcal{F}}_n^s$):}} They are derived from the QoS log matrix. We perform non-negative matrix decomposition \cite{nmf} of ${\mathcal{Q}}$ to obtain the collaborative QoS features of $u_i$ and $s_j$, each with dimension $d_n$.
    
    \noindent
    {\emph{(iii) Similarity features (${\mathcal{F}}_s^u, {\mathcal{F}}_s^s$):}} They are also extracted from the QoS log matrix. We employ the cosine similarity metric (CSM) to obtain the similarity between pairwise users $(u_i, u_k)$ and services $(s_j, s_k)$, computed as: 
    
    \begin{equation}
        CSM(u_i, u_k) =   \dfrac {{\mathcal{Q}}(i) \cdot {\mathcal{Q}}(k)} {\left\| {\mathcal{Q}}(i) \right\| _{2}\left\| {\mathcal{Q}}(k)\right\| _{2}}
    \end{equation}
    \begin{equation}
        CSM (s_j, s_k) =   \dfrac {{\mathcal{Q}}^T(j) \cdot {\mathcal{Q}}^T(k)} {\left\| {\mathcal{Q}}^T(j) \right\| _{2}\left\| {\mathcal{Q}}^T(k)\right\| _{2}}
    \end{equation}
    
    \noindent
    It is worth noting that the similarity between two users/services is 0 if they do not have any common invocations. Therefore, in general, the similarity features remain sparse due to the sparsity of ${\mathcal{Q}}$. Furthermore, depending on the number of users and services, the similarity feature vector may be a high-dimensional vector. It may be noted that the length of the similarity feature vector for each user $u_i$ and service $s_j$ are $n$ and $m$, respectively. Therefore, we employ two different auto-encoders \cite{STACKED-AUTOENCODER} for the users and services separately to obtain the low-dimensional similarity feature vector of length $d_s$.

    \noindent
    {\textbf{Contextual features (${\mathcal{F}}_c^u, {\mathcal{F}}_c^s$):}} We utilize the contextual information associated with each $u_i$ and $s_j$ to prepare the contextual features. The user contextual features include the user ID, region, and autonomous system (AS) information, whereas the service contextual features comprise the service ID, service region, and service provider information.     
    We use the one-hot encoding for each contextual information. Here again, the dimension of the feature vector becomes large due to the high number of users and services. Therefore, we employ auto-encoders to obtain the contextual feature vector of length $d_c$. 
        
    Finally, all the above feature vectors are concatenated to construct the feature embedding ${\mathcal{E}}^{i}_{1} = ({\mathcal{F}}_t^{<u, i>} || {\mathcal{F}}_n^{<u, i>} || {\mathcal{F}}_s^{<u, i>} || {\mathcal{F}}_c^{<u, i>}) \in {\mathcal{E}}_1$ and ${\mathcal{E}}^{j}_{2} = ({\mathcal{F}}_t^{<s, j>} || {\mathcal{F}}_n^{<s, j>} || {\mathcal{F}}_s^{<s, j>} || {\mathcal{F}}_c^{<s, j>}) \in {\mathcal{E}}_2$ for each user $u_i$ and service $s_j$, respectively. Here, ${\mathcal{F}}_t^{<u, i>}$ refers to the statistical features for $u_i$. The rest of them are denoted similarly.  
    The length of the initial feature embedding (say, $f$) for each user/service is, therefore, $(5 + d_n + d_s + d_c)$. The initial feature embedding matrix $F^0$ is defined in Eq. \ref{eq:embed}.
    \begin{equation} \label{eq:embed}
        F^0 = (\rho^0_{ij}) \in \mathbb{Z}^{N \times f};
        \rho^0_{i} = 
        \begin{cases}
          {\mathcal{E}}^{i}_{1} \in {\mathbb{R}}^f, \text{ if } i \le n\\
          {\mathcal{E}}^{i - n}_{2} \in {\mathbb{R}}^f ,~ \text{otherwise}
        \end{cases}
    \end{equation}
    \noindent
    The contextual features are less proficient than QoS features in capturing the intricate relationship among users and services, particularly in the context of QoS prediction. However, relying solely on QoS features often results in lower accuracy due to the sparse nature of the QoS log matrix. Therefore, in this paper, we combine both types of features to create the initial feature embedding.     
    In addition to the initial feature embedding, we incorporate spatial collaborative features obtained from the MhGCMF, which forms the core component of SORRQP. 
    The graph architecture in the MhGCMF enables each user/service to integrate the neighbor information into its embedding, resulting in richer representations that facilitate effective QoS prediction. We now discuss the details of MhGCMF. The primary goal of the MhGMCF is to learn the spatial features from the initial feature embedding of the users and services, with the aim of enhancing the efficiency of QoS prediction. The GCMF Unit (GCMFU) is the central component for MhGCMF. Therefore, we begin with explaining the GCMFU.

 %%%%%%%%%%%%%%%%%%%%%%%%%%%%%%%%%%%%%%%%%%%%%%%%%%%%%%%%%%%%
 
    \subsubsection{Architecture of GCMFU} 
    GCMFU aggregates the neighborhood features of a node $v_i^k \in (V_1 \cup V_2),~ k\in \{1, 2\}$ from its neighbors, as modeled by the two equations shown in Fig. \ref{fig:mhgcn}(a). 
    Here, we leverage the adjacency matrix representation of $\mathcal{G}$ to aggregate the neighborhood features. To preserve the self features of each node in the aggregation, we additionally make the diagonal entries of the adjacency matrix as 1. The modified adjacency matrix $\mathcal{A} \in \mathbb{R}^{N \times N}$ is defined in Eq. \ref{eq:adj}.
    \noindent
    \begin{equation}\label{eq:adj}
    \small
        \mathcal{A} = (a_{ij}) \in \{0, 1\}^{N \times N};
        a_{ij} = 
        \begin{cases}
          1 ,~ \text{if } i~=~j \text{ or } e_{ij} \in E \\
          0 ,~ \text{otherwise}
        \end{cases} 
    \end{equation}
    \noindent
    Where, $N = n + m$. We then normalize $\mathcal{A}$, as shown in Eq. \ref{eq:norm_adj}, to avoid the scaling discrepancy caused by the varying degrees of the nodes in $\mathcal{G}$ because of the sparsity of $\mathcal{Q}$ and generate $\bar{\mathcal{A}}$.
    The diagonal degree matrix ($\mathcal{D}$), as defined in Eq. \ref{eq:diag}, is used for the normalization. The normalization reduces the impact of the higher degree nodes in QoS prediction.
    \begin{equation} \label{eq:norm_adj}
    \small
    % \centering
    % \begin{split}
        \bar{\mathcal{A}} = \mathcal{D}^{-1/2} ~ \mathcal{A} ~ \mathcal{D}^{-1/2} \\
    \end{equation}
    % where,  
    \begin{equation} \label{eq:diag}
    \mathcal{D} = (d_{ij}) \in \mathbb{Z}^{N \times N};~~
        d_{ij} = 
        \begin{cases}
          \sum \limits_{k=1}^N {\mathcal{A}} (i, k), \text{ if } i = j\\
          0 ,~ \text{otherwise}
        \end{cases}
    % \end{split}
    \end{equation}

\noindent
The GCMFU receives $\bar{\mathcal{A}}$ and a feature embedding matrix (say, $F^j$) as its input. It then performs a non-linear transformation on $F^j$ with the help of a set of learnable parameters (refer to Fig. \ref{fig:mhgcn}(a)) and produces the updated feature embedding matrix (say, $F^{j+1}$) by incorporating the spatial information of the neighborhood nodes. We now discuss the detailed architecture of the MhGCMF. 

%%%%%%%%%%%%%%%%%%%%%%%%%%%%%%%%%%%%%%%%%%%%%%%%%%%%%%%%%%%%%%%%%%%%%%%%%%%%%

\subsubsection{Architecture of MhGCMF} 
Fig. \ref{fig:mhgcn}(b) summarizes the architecture of the MhGCMF. Given the normalized adjacency matrix $\bar{\mathcal{A}}$ and the initial feature embedding matrix $F^0$, the objective of the MhGCMF is to learn the spatial collaborative features for each user/service and predict the QoS value of a given user-service pair.

In MhGCMF, $F^0$ first undergoes an initial transformation by passing through a dense layer, resulting in the transformed feature embedding $F^1$. 
Subsequently, $F^1$ and $\bar{\mathcal{A}}$ are directed to the multi-head GCMFU (i.e., MhGCMFU) block. MhGCMFU comprises $N_h$ number of GCMFUs followed by a $1 \times 1$ convolution layer. The ``multi-head" in MhGCMFU refers to the presence of multiple GCMFUs.
The outputs of these multiple GCMFUs are then concatenated channel-wise, and the resulting tensor passes through a $1 \times 1$ convolutional layer. This entire process is repeated $t$ times, where the output of MhGCMFU is fed back into itself.
The output generated by each MhGCMFU block is once again concatenated channel-wise and then forwarded to another $1 \times 1$ convolutional layer. For each $1 \times 1$ convolutional layer, there is a corresponding tuple indicating (padding, stride, and number of filters), as illustrated in Fig. \ref{fig:mhgcn}(b).
The output of the final $1 \times 1$ convolutional layer is split row-wise to produce the user and service embedding matrices. These matrices are subsequently multiplied to generate the predicted QoS log matrix. 
We train the MhGCMF for each user-service pair $(u_i, s_j) \in \mathcal{U} \times \mathcal{S}$, such that $q_{ij} \ne 0$ using Cauchy Loss Function, as shown in Eq. \ref{eq:cauchy_loss}.
\begin{equation} \label{eq:cauchy_loss}
\small 
   \mathcal{C_L} = \sum_{q_{ij} \ne 0}^{} \ln{(1+ \frac{|| q_{ij} - \hat{q}_{ij}|| ^2}{\gamma^2})}
\end{equation}
where $\gamma$ is a hyper-parameter that can be tuned externally, and $\hat{q}_{ij}$ is the predicted QoS value. 
The outlier resilience characteristics of the Cauchy loss function ensure that SORRQP also exhibits outlier resilience.
However, SORRQP alone cannot address the other issues, such as grey sheep and cold start. In the next section, we address the issue of grey sheep. 

%%%%%%%%%%%%%%%%%%%%%%%%%%%%%%%%%%%%%%%%%%%%%%%%%%%%%%%%%%%%%%%%%%%%%%%%%
\subsection{Grey sheep Detection Block (GD)} 
Collaborative Filtering (CF) highly relies on the collaborative relationship between users and services and operates on the premise that similar users and services exist within the system.
However, it is worth noting that there can be a small group of users or services with QoS patterns that deviate from the norm, making them distinct from the majority. These users and services are often referred to as {\emph{grey sheep}}, representing an anomalous subset within the user or service community. The existence of grey sheep users or services may occasionally lead to substantial inaccuracies in QoS predictions. In this context, our approach begins with addressing the task of identifying grey sheep Users/Services (GSU/GSS). Subsequently, we tackle the challenge of QoS prediction for these identified GSUs and/or GSSs. The GD block within ARRQP primarily focuses on identifying the instances of grey sheep within the QoS data.

To capture the distinct QoS pattern exhibited by a user/service, here, we introduce two concepts: {\emph{reliability score}} and {\emph{Grey sheep Anomaly (GA) score}} \cite{GS} for each user/service. These two scores together aid in determining whether a particular user/service qualifies as a GSU/GSS.

\begin{definition}[Reliability Score]\label{def:reliability}
The reliability score of a user $u_i \in \mathcal{U}$ and service $s_j \in {\mathcal{S}}$, denoted by $\mathscr{R}(u_i)$ and $\mathscr{R}(s_j)$, respectively, are defined as:

\begin{minipage}{0.22\textwidth}
\begin{equation}\label{eq:r_score_user}\scriptsize
    \begin{aligned}
        & {\mathscr{R}}(u_i) = \\
        & 1 - \left(\frac{\sigma^u_i - \min\limits_{u_k \in {\mathcal{U}}} (\sigma^u_k)} {\max\limits_{u_k \in {\mathcal{U}}} (\sigma^u_k) - \min\limits_{u_k \in {\mathcal{U}}} (\sigma^u_k)}\right);
    \end{aligned}
\end{equation}
\end{minipage}
\begin{minipage}{0.22\textwidth}\scriptsize
\begin{equation}\label{eq:r_score_service}
    \begin{aligned}
        & {\mathscr{R}}(s_j) = \\
        & 1 - \left(\frac{\sigma^s_j - \min\limits_{s_k \in {\mathcal{S}}} (\sigma^s_k)} {\max\limits_{s_k \in {\mathcal{S}}} (\sigma^s_k) - \min\limits_{s_k \in {\mathcal{S}}} (\sigma^s_k)}\right)
    \end{aligned}
\end{equation}
\end{minipage}
\noindent
 $\sigma^u_i$ and $\sigma^s_j$ represent the standard deviation of the QIVs of $u_i$ and $s_j$, respectively (refer to Table \ref{tab:feature_vector}).
\hfill $\blacksquare$
\end{definition}

\noindent
It may be noted from Eq.s \ref{eq:r_score_user} and \ref{eq:r_score_service} that a user/service is said to be more reliable than another if the standard deviation of the QIV of the former is lower than that of the later. We now define GA score in Definition \ref{def:abnormality}.

\begin{definition}[Grey sheep Anomaly (GA) Score]\label{def:abnormality}
The GA score of a user $u_i \in \mathcal{U}$ and service $s_j \in {\mathcal{S}}$, denoted by $\mathscr{G}(u_i)$ and $\mathscr{G}(s_j)$, respectively, are defined as:

% \begin{equation}\label{eq:ab_score_user}\scriptsize
%     \mathscr{G}(u_i) = \frac{\sum \limits_{s_j \in {\mathcal{S}}_i}\Bigl(|q_{ij} - \mu^u_i - \bar{\mu}^s_j| \times {\mathscr{R}}(s_j)\Bigr)}{|{\mathcal{S}}_i|}
% \end{equation}
% \begin{equation}\label{eq:ab_score_service}\scriptsize
%     \mathscr{G}(s_j) = \frac{\sum \limits_{u_i \in {\mathcal{U}}_j}\Bigl(|q_{ij} - \mu^s_j - \bar{\mu}^u_i| \times {\mathscr{R}}(u_i)\Bigr)}{|{\mathcal{U}}_j|}
% \end{equation}
% \begin{equation}
% \scriptsize
% \bar{\mu}^s_j = \frac{\left(\sum \limits_{u_i \in {\mathcal{U}}_j} q_{ij}\right) - \max \limits_{u_i \in {\mathcal{U}}_j}(q_{ij}) - \min\limits_{u_i \in {\mathcal{U}}_j}(q_{ij})}{|{\mathcal{U}}_j| - 2}
% \end{equation}
\begin{equation}\label{eq:ab_score_user}\scriptsize
    \mathscr{G}(u_i) = \left({\sum \limits_{s_j \in {\mathcal{S}}_i}\Bigl(|q_{ij} - \mu^u_i - \bar{\mu}^s_j| \times {\mathscr{R}}(s_j)\Bigr)}\right)/{|{\mathcal{S}}_i|}
\end{equation}
\begin{equation}\label{eq:ab_score_service}\scriptsize
    \mathscr{G}(s_j) = \left({\sum \limits_{u_i \in {\mathcal{U}}_j}\Bigl(|q_{ij} - \mu^s_j - \bar{\mu}^u_i| \times {\mathscr{R}}(u_i)\Bigr)}\right)/{|{\mathcal{U}}_j|}
\end{equation}
\begin{equation}
\scriptsize
\bar{\mu}^s_j = \left({\left(\sum \limits_{u_i \in {\mathcal{U}}_j} q_{ij}\right) - \max \limits_{u_i \in {\mathcal{U}}_j}(q_{ij}) - \min\limits_{u_i \in {\mathcal{U}}_j}(q_{ij})}\right)/\left({|{\mathcal{U}}_j| - 2}\right)
\end{equation}
\noindent
$\bar{\mu}^u_i$ is also defined similarly.
${\mathcal{S}}_i$ be the set of services invoked by $u_i$. ${\mathcal{U}}_j$ be the set of users invoked $s_j$. $\mu^u_i$ and $\mu^s_j$ represent the mean of the QIVs of $u_i$ and $s_j$, respectively. 
\hfill $\blacksquare$
\end{definition}
\noindent
It may be noted from the above equations that the GA score of each user $u_i$ (service $s_j$) is computed over its respective QIV, denoted by ${\mathcal{Q}}(i)$ (${\mathcal{Q}}^T(j)$). For each invocation $q_{ij}$ ($\ne 0$) made by $u_i$ (for $s_j$), we compute the deviation with respect to the mean of QIV of $u_i$ ($s_j$) and centralized mean of $s_j$ ($u_i$). 
The average of these weighted deviations, computed over the QIV of $u_i$ ($s_j$), is referred to as the GA score of $u_i$ ($s_j$). 
The reliability score of the corresponding service is used as the weight in this computation. Notably, if a service has a high reliability score, the corresponding deviation carries more weight than one associated with a service with a lower-reliability score. It may be noted the GA score of a user only depends on the set of services it invoked. Therefore, the GA score is not affected by the data sparsity. 

We now define the Grey sheep user (GSU) and service (GSS) below.

\begin{definition}[Grey sheep user (GSU)]
A user $u_i \in {\cal{U}}$ (service $s_j \in {\cal{S}}$) is called GSU (GSS), if ${\mathscr{G}}(u_i)$ (${\mathscr{G}}(s_j)$) is more than a given threshold ${\tau}_{\mathscr{G}}^u$ (${\tau}_{\mathscr{G}}^s$). 
\hfill $\blacksquare$
\end{definition}

\noindent
${\tau}_{\mathscr{G}}^u$ and ${\tau}_{\mathscr{G}}^s$ are hyper-parameters. 
In this paper, we consider ${\tau}_{\mathscr{G}}^u = \mu_{\mathscr{G}}^u + c * \sigma_{\mathscr{G}}^u$ (${\tau}_{\mathscr{G}}^s = \mu_{\mathscr{G}}^s + c * \sigma_{\mathscr{G}}^s$), where $\mu_{\mathscr{G}}^u$ and $\sigma_{\mathscr{G}}^u$ ($\mu_{\mathscr{G}}^s$ and $\sigma_{\mathscr{G}}^s$) are the mean and standard deviation of the GA scores of users (services), respectively. $c$ is a hyper-parameter, which can be tuned externally.

Once the GD block identifies the GSU/GSS, the next step involves predicting the QoS values for these identified GSUs/GSSs using our next prediction block, GRRQP. In the following section, we discuss the specifics of the GRRQP.

\subsection{GRRQP Block}
The GRRQP is primarily designed to provide QoS predictions for GSS/GSU. The rationale behind introducing the GRRQP block stems from the recognition that grey sheep users/services exhibit markedly distinct QoS patterns compared to the majority of users/services. Consequently, relying extensively on collaborative features, as employed in SORRQP, may not be adequate for achieving higher QoS predictions for GSU/GSS. Furthermore, the prediction of QoS values for GSU/GSS in conjunction with other users/services may not be advisable, given that GSU/GSS possess distinct characteristics that set them apart from the rest. Therefore, here, we design separate architectures for the QoS prediction tailored specifically for GSU/GSS. 

The GRRQP is an improvisation of SORRQP, featuring supplementary MLPs (Multi-Layer Perceptrons) in addition to the MhGCMF model discussed previously. Here, we have three MLPs designed specifically for (a) non-grey sheep users and GSSs, (b) GSUs and non-grey sheep services, and (c) GSUs and GSSs. 
The feature vector for each MLP is constructed by concatenating the user features and the service features. If the user $u_i$ is a non-grey sheep user, the embedding vector $E^u_i$ (i.e., $i^{th}$ row of $E^u$) obtained from MhGCMF is used as its features. However, if $u_i$ is a grey sheep user, its feature is generated by concatenating the following components: (i) the embedding vector $E^u_i$ obtained from MhGCMF, (ii) the initial feature embedding vector ${\mathcal{E}}_1^i$, (iii) the GA score ${\mathscr{G}}(u_i)$, and (iv) the number of invocations of $u_i$.
The service features are also generated similarly. The MLPs are trained for the user-service pair in the respective categories for which $q_{ij} \ne 0$.

In the following subsection, we illustrate our final prediction block, designed to handle the cold start issue. 
%%%%%%%%%%%%%%%%%%%%%%%%%%%%%%%%%%%%%%%%%%%%%%%%%%%%%%%%%%%%%%%%%%
\subsection{CRRQP Block}
The term {\emph{cold start}} pertains to a scenario in which a new user/service is introduced into a system. This situation poses a significant challenge for QoS 
 prediction due to the lack of available data. The SORRQP system is ill-suited to handle the cold start scenario for the following reasons:
(a) The similarity and statistical features in the initial feature embedding cannot provide meaningful information for the newly added user/service due to the absence of historical data.
(b) The newly introduced user/service forms an isolated node within the QIG, indicating no interactions among other users/services. These isolated nodes lack the ability to capture spatial features effectively. As a result, this limitation poses a challenge for SORRQP in making accurate predictions under such circumstances. Therefore, we propose another prediction block, CRRQP, to address the above issue. CRRQP is an enhancement over SORRQP.

In addition to MhGCMF, CRRQP is also equipped with three MLPs to predict the QoS of the following target user-service pairs. 
(a) Cold Start User (CSU): Here, the target user is new. However, the service has sufficient past data.
(b) Cold Start Service (CSS): In this case, the target service is new, while the user has historical data.
(a) Cold Start user/service Both (CSB): This scenario involves both the user and service being newly introduced entities without any historical data available for either of them.

The three MLPs are trained separately with three different sets of input features. The input features are constructed by concatenating the user and the service features. If the user/service $u_i$/$s_j$ is not a newly added, the embedding vector $E^u_i$/$E^s_j$ obtained from MhGCMF is used as its features. However, if $u_i$/$s_j$ is newly added, its feature is generated by concatenating its contextual features and the collaborative features obtained using non-negative matrix decomposition.

In the next subsection, we discuss our final block for outlier detection.  
%%%%%%%%%%%%%%%%%%%%%%%%%%%%%%%%%%%%%%%%%%%%%%%%%%%%%%%%%%%%%%%%%%%
\subsection{Outlier Detection Block}\label{subsec:anomaly_detection}
In this paper, we employ an unsupervised learning algorithm based on the isolation forest ($i$Forest) algorithm \cite{iforest} to detect the outliers from scratch. $i$Forest algorithm explicitly isolates the outliers rather than profiling the inliers converse to the distance or density-based approaches for the outliers detection. The strength of the $i$Forest approach is its effectiveness over different benchmarks and high scalability. It uses an ensemble of isolation trees ($i$Tree), whose branch grows iteratively; this way, it isolates every single data point. Finally, in each $i$Tree, the outliers are instances that are isolated near the root of the tree and have comparatively short average path lengths, consecutively, inliers isolated to the far end of the $i$Tree. After identifying the outliers, we proceed to eliminate them from the dataset, with the ratio of removed outliers denoted by $\lambda$ (e.g., $\lambda = 0.1$ refers to 10\% of the total dataset considered as the outlier). This step is essential to evaluate the performance of our framework without outliers.

\begin{figure}
    \centering
    \includegraphics[width=\linewidth]{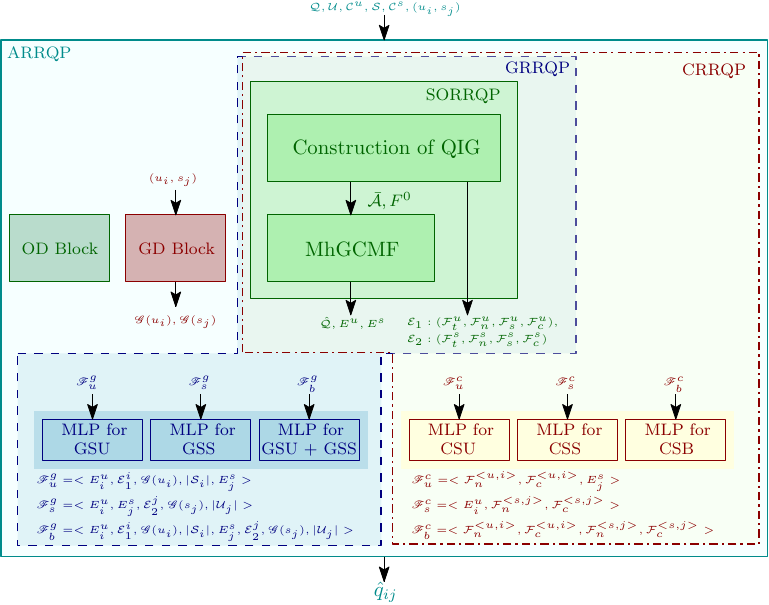}
    \caption{Overall architecture ARRQP}
    \label{fig:flow}
\end{figure}

Fig. \ref{fig:flow} shows the overall architecture of ARRQP. In the next section, we analyze the performance of ARRQP.

% with default parameters (where the number of trees and the sub-sampling size are 100 and 256, respectively) 

\section{Experimental Results}\label{sec:result}
\noindent
In this section, we present the experimental analysis of our framework. We implemented our proposed framework in TensorFlow 2.6.2 with Python 3.6.9. All experiments were executed on the Linux-5.4.0-133-generic-x86\_64-with-Ubuntu-18.04-bionic with Intel(R) Core(TM) i9-10885H CPU @ 2.40GHz x86\_64 architecture, 16 Core(s), 128 GB RAM, with the following cache L1d: 32K, L1i:32K, L2:256K, L3:16384K.

\subsection{Experimental Setup}
In this subsection, we present a comprehensive overview of the datasets, performance metrics, model configurations utilized to conduct our experiment, and finally, the experimental analysis. We begin by detailing the datasets used in our study.

\subsubsection{Datasets}
\noindent
We used the benchmark and publicly available WS-DREAM \cite{WSDREAM} datasets. The dataset contains two different QoS parameters: response time (RT) and throughput (TP). Table \ref{tab:dataset_table} comprehensively provides the details of the dataset.

% The WS-DREAM-1 dataset not only provides a QoS invocation log but also contains the contextual details of users and services. Table \ref{tab:dataset_table} provides the statistical measures such as minimum value, maximum value, mean, median, and standard deviation for each of the two QoS parameters along with contextual information of users and services that comprises the number of users, number of services, number of services invocations by the users, number of users/services region, number of users AS, number of services region, and number of service providers.

\begin{table}[!b]
    \centering
    \caption{WS-DREAM datasets details for RT and TP}
    \begin{tabular}{c|c|c}
         \hline
        Statistics & RT & TP \\\hline \hline
        Number of Users & \multicolumn{2}{c}{$339$}\\ \hline
        Number of User's Regions & \multicolumn{2}{c}{$31$} \\\hline
        Number of User's AS & \multicolumn{2}{c}{$137$} \\\hline
        Number of Services & \multicolumn{2}{c}{$5825$} \\\hline
        Number of Service's Regions & \multicolumn{2}{c}{$74$} \\\hline
        Number of Service's Providers & \multicolumn{2}{c}{$2699$} \\\hline
        Number of Valid Invocations & $1873838$ & $1831253$ \\\hline
         % Minimum & 0.0010 & 0.0040 \\ \hline
         Range (min-max) & $0.0010-19.9990$ & $0.0040-1000$ \\\hline
         % Maximum &	19.9990 & 1000 \\ \hline
         Mean &	$0.9086$ & $47.5617$ \\ \hline
         Median	& $0.3200$ & $13.9530$ \\ \hline
         Standard Deviation & $1.9727$ & $110.7970$ \\ \hline
    \end{tabular}
    \label{tab:dataset_table}
\end{table}

\begin{table*}[!t]
    \centering
     \scriptsize 
    \caption{Performance of ARRQP on Response Time (RT)}
    \begin{tabular}{l|l|c|c|c|c|c|c|c|c}
    \hline
     % &  &   \\ \cline{3-10}
    Anomaly   & \multirow{2}{*}{Methods} & \multicolumn{4}{c|}{MAE} & \multicolumn{4}{c}{RMSE} \\ \cline{3-10}
    Addressed & & RT-5 & RT-10 & RT-15 & RT-20 & RT-5 & RT-10 & RT-15 & RT-20 \\\hline\hline 
    \multirow{5}{*}{None} & UPCC \cite{upcc} & 0.7695 & 0.7201 & 0.6521 & \textcolor{blue}{0.5921} & 1.6940 & 1.6101 & 1.4959 & 1.3993 \\ 
    & IPCC \cite{ipcc} &	0.7326 &	0.7125	& 0.6783 &	0.6503 &	1.6346 &	1.5439 &	1.4104	& \textcolor{blue}{1.3109} \\ 
    & WSRec \cite{wsrec} &	\textcolor{blue}{0.6794}	& \textcolor{blue}{0.6211} &	\textcolor{blue}{0.6037} &	0.6020 &	\textcolor{blue}{1.4884} &	\textcolor{blue}{1.4265} &	\textcolor{blue}{1.3684} &	1.3515 \\  \cline{2-10}
    & $\mathcal{I}$ (\%) & \textbf{46.46} &	\textbf{49.49} &	\textbf{54.16} &	\textbf{56.32} &	\textbf{16.83} &	\textbf{19.69} &	\textbf{20.51} &	\textbf{23.41}  \\ \hline

    % \multirow{2}{*}{grey sheep}
    % & RAP \cite{rap} & 0.5641	& 0.4914	& 0.4639	& 0.4531 & 	1.4346 & 	1.2783 & 	1.215	& 1.1631 \\ 
    % & $\mathcal{I}$ (\%) & & & & & & & &  \\ \hline

    \multirow{20}{*}{${\mathds{S}} + {\mathds{C}}$}
    % &  UIPCC/ \cite{uipcc} & 0.6234 & 0.5365 & 0.4965 & 0.5982 & 1.4078 & 1.3043 & 1.2467 & 1.2012 \\ 
    &  SN-MF \cite{snmf} & 0.6318 & 0.5377 & 0.5008 & 0.4871 & 1.3968 & 1.2695 & 1.2264 & 1.2073\\ 
    &  LACF \cite{lacf} & 0.6299 & 0.5601 & 0.5102 & 0.4774 & 1.4399 & 1.3377 & 1.2694 & 1.2220\\ 
    &  NMF \cite{nmf} & 0.6182 & 0.6040 & 0.5990 & 0.5982 & 1.5746 & 1.5494 & 1.5345 & 1.5531 \\
    &  LFM \cite{lfm} & 0.5783 & 0.5644 & 0.5438 & 0.5350 & 1.5010 & 1.3201 & 1.2719 & 1.2269 \\
    &  LE-MF \cite{lemf} & 0.5734 & 0.5136 & 0.4827 & 0.4646 & 1.3703 & 1.2511 & 1.2041 & 1.1809 \\
    &  PMF \cite{pmf} & 0.5678 & 0.4996 & 0.4720 & 0.4492 & 1.4735 & 1.2866 & 1.2163 & 1.1828 \\
    & NIMF \cite{nimf} & 0.5514 & 0.4884 & 0.4534 & 0.4357 & 1.4075 & 1.2745 & 1.1980 & 1.1678\\
    & LN-LFM \cite{lnlfm} & 0.5501 & 0.4889 & 0.4611 & 0.4424 & 1.3145 & 1.2368 & 1.2073 & 1.1568\\ 
    & RSNMF \cite{rsnmf} & 0.5438 & 0.4868 & 0.4492 & 0.4371 & 1.4032 & 1.2689 & 1.2067 & 1.1568\\
    & LBR \cite{lbr} & 0.5389 & 0.5292 & 0.5180 & 0.4941 & 1.4130 & 1.3481 & 1.3260 & 1.3147 \\ 
    & NAMF \cite{namf} & 0.5384 & 0.4850 & 0.4529 & 0.4350 & 1.3850 & 1.2592 & 1.2071 & 1.1443 \\ 
    & GeoMF \cite{geomf} & 0.5305 & 0.4827 & 0.4495 & 0.4366 & 1.3152 & 1.2190 & 1.1742 & 1.1528\\
    & CNMF \cite{cnmf} & 0.5289 &	0.4713 & 0.4316 & 0.4136 & 1.3053 &	1.2373 & 1.1360	& 1.1161 \\
    & CSMF \cite{csmf} & 0.5286 & 0.4557 & 0.4225 & 0.4180 & 1.3473 & 1.2315 & 1.1780 & 1.1210\\
    & LMF-PP \cite{lmfpp} & 0.5285 & 0.4725 & 0.4472 & 0.4260 & 1.3410 & 1.2419 & 1.2102 & 1.1625\\
    & JCF \cite{jcf} & 0.5132 & 0.4665 & 0.4504 & 0.4365 & 1.3328 & 1.2505 & 1.1854 & 1.1802\\
    & NDMF \cite{ndmf} & 0.4880 & 0.4304 & 0.3845 & 0.3665 & 1.3495 & 1.2349 & 1.1569 & 1.1294 \\
    & EFMPred \cite{efm} & 0.4460 & 0.3750 & 0.3630 & 0.3480 & 1.3530 & 1.2810 & 1.2530 & 1.2360\\ 
    & MSDAE \cite{msdae} & 0.4267 & \textcolor{blue}{0.3640} & \textcolor{blue}{0.3434} & 0.3275 & 1.2853 & 1.2436 & 1.1695 & 1.1283\\
    & LAFIL \cite{lafil} & \textcolor{blue}{0.3880} & 0.3664 & 0.3460 & \textcolor{blue}{0.3268} & \textcolor{blue}{1.2674} & \textcolor{blue}{1.1926} & \textcolor{blue}{1.1106} & \textcolor{blue}{1.0923} \\\cline{2-10}
    & $\mathcal{I}$ (\%) &  \textbf{6.26} & \textbf{13.82} & \textbf{19.42} & \textbf{22.58} & \textbf{2.32} & \textbf{3.94} & \textbf{2.06} & \textbf{8.95}  \\ \hline

    \multirow{6}{*}{${\mathds{S}} + {\mathds{C}}  + {\mathds{O}}$}
    & DCALF \cite{dcalf} & 0.5127 & 0.4544 & 0.4346 & 0.4246 & 1.3731 & 1.2450 & 1.2001 & 1.1759  \\   
    & NCF \cite{ncf} & 0.4400 & 0.3850 & 0.3720 & 0.3620 & 1.3250 & 1.2830 & 1.2530 & 1.2050 \\ 
    &  SPP+LLMF \cite{llmf} & 0.4350	& 0.3700 & 	0.3550 & 0.3360 & 1.3340 & 	1.2500	& 1.1920 & 1.1740 \\ 
    & DNM \cite{dnmm} & 0.4125 & 0.3726 & 0.3678 & 0.3575 & 1.3463 & 1.2673 & 1.2490 & 1.2298 \\
    & LDCF \cite{ldcf} & 0.4020 & 0.3670 & 0.3450 & 0.3310 & 1.2770 & 1.2330 & 1.1690 & 1.1380 \\
    & DCLG \cite{dclg} & 0.3850 & 0.3460 & 0.3260 & 0.3140 & 1.2630 & 1.1850 & 1.1430 & 1.1200 \\
    & HSA-Net \cite{hsanet} & \textcolor{blue}{0.3670} & \textcolor{blue}{0.3270} & \textcolor{blue}{0.3070} & \textcolor{blue}{0.2950} & \textcolor{blue}{1.2490} & \textcolor{blue}{1.1660} & \textcolor{blue}{1.1260} & \textcolor{blue}{1.0910} \\\cline{2-10}
    & $\mathcal{I}$ (\%) &  \textbf{0.90} & \textbf{4.06} & \textbf{9.86} & \textbf{14.24} & \textbf{0.89} & \textbf{1.75} & \textbf{3.40} & \textbf{8.85} \\ \hline

    % \multirow{3}{*}{Sparsity + grey sheep + Cold Start}
    % & LRMF \cite{lrmf} & 0.5516 & 0.4785 & 0.4460 & 0.4340 & 1.4151 &	1.2576 & 1.2128 & 1.1422 \\
    % & $\mathcal{I}$ (\%) & & & & & & & &  \\ \hline
    
    \multirow{1}{*}{${\mathds{S}} + {\mathds{GS}}  + {\mathds{O}} + {\mathds{C}}$}  
    & \textbf{ARRQP} & \textbf{0.3637} & \textbf{0.3137} & \textbf{0.2767} & \textbf{0.2530} & \textbf{1.2379} & \textbf{1.1456} & \textbf{1.0877} & \textbf{0.9945} \\\hline

    \multicolumn{10}{r}{*  Improvement of ARRQP over the second-best method highlighted in \textcolor{blue}{blue}}
    \end{tabular}
    
    \label{tab:rt_soa}
\end{table*}

\begin{table*}[!t]
    \scriptsize 
    \centering
    \caption{Performance of ARRQP on Throughput (TP)}
    \begin{tabular}{l|l|c|c|c|c|c|c|c|c}
    \hline
    Anomaly   & \multirow{2}{*}{Methods} & \multicolumn{4}{c|}{MAE} & \multicolumn{4}{c}{RMSE} \\ \cline{3-10}
    Addressed & & TP-5 & TP-10 & TP-15 & TP-20 & TP-5 & TP-10 & TP-15 & TP-20 \\\hline\hline
    \multirow{4}{*}{None} 
     & UPCC  &	27.5760 &	23.0130 &	20.6800 &	19.2940	& 63.7540 &	56.6210 &	51.9550 &	47.6510 \\
     & IPCC  &	26.8640 &	26.0590 &	25.8790 &	24.2610 & 65.7900 &	61.1320 &	59.1500 &	54.6970 \\
     & WSRec &	\textcolor{blue}{26.7330} &	\textcolor{blue}{22.6940} &	\textcolor{blue}{20.5470} &	\textcolor{blue}{18.9640} & \textcolor{blue}{63.4980} &	\textcolor{blue}{56.4590} &	\textcolor{blue}{51.7890} &	\textcolor{blue}{46.9510} \\\cline{2-10}
    & $\mathcal{I}$ (\%) & \textbf{50.40} & \textbf{52.53} & 	\textbf{52.50} &	\textbf{53.97} &\textbf{36.53} &	\textbf{34.39} &	\textbf{32.48} & \textbf{29.73} \\ \hline

    \multirow{20}{*}{${\mathds{S}} + {\mathds{C}}$}
    & CSMF \cite{csmf} & 28.3060 & 26.5850 & 25.6250 & 21.7420 & 72.7040 & 70.0210 & 68.6180 & 61.6790 \\
    & LBR \cite{lbr} & 26.1610 & 21.0060 & 18.5840 & 16.5510 & 68.0310 & 57.7280 & 54.4410 & 51.6790 \\
    & NMF \cite{nmf} & 25.7529 & 17.8411 & 15.8939 & 15.2516 & 65.8517 & 53.9896 & 51.7322 & 48.6330 \\
    & GeoMF \cite{geomf} & 24.7465 & 22.4728 & 17.7908 & 16.2852 & 57.7842 & 49.2456 & 45.3255 & 43.9545 \\
    & LACF \cite{lacf} & 22.9737 & 19.4489 & 17.5887 & 16.4580 & 58.7857 & 52.9270 & 49.5650 & 47.4108 \\
    & RSNMF \cite{rsnmf} & 21.4302 & 17.2305 & 14.6880 & 14.3654 & 60.7994 & 50.5298 & 45.2647 & 43.5822 \\
    & PMF \cite{pmf} & 19.9034 & 16.1755 & 15.0956 & 14.6694 & 54.0508 & 46.4439 & 43.7957 & 42.4855 \\
    & LB-FM \cite{lbfm} & 19.8460 & 17.4610 & 17.0460 & 15.8390 & 73.8420 & 65.6800 & 67.0930 & 61.8030 \\
    & LRMF \cite{lrmf} & 19.1090 & 15.9494 & 14.5974 & 13.9206 & 58.0719 & 48.2718 & 44.0682 & 41.7880 \\
    & LN-LFM \cite{lnlfm} & 18.6512 & 16.0634 & 14.7664 & 14.1264 & 52.4326 & 46.8756 & 44.3871 & 43.0892 \\
    & LMF-PP \cite{lmfpp} & 18.3091 & 15.9125 & 14.7450 & 14.1033 & 51.7765 & 46.1418 & 42.9927 & 41.4084 \\
    & NAMF \cite{namf} & 18.0836 & 15.9808 & 14.6661 & 13.9386 & 52.8658 & 44.0788 & 43.0206 & 40.7481 \\
    & NIMF \cite{nimf} & 17.9297 & 16.0542 & 14.4363 & 13.7099 & 51.6573 & 45.9409 & 43.1596 & 41.1689 \\
    & NDMF \cite{ndmf} & 16.3818 & 13.9317 & 12.5043 & 11.7204 & 50.9612 & 43.9095 & 42.5319 & 39.9431 \\
    & LAFIL \cite{lafil} & \textcolor{blue}{13.9875} & \textcolor{blue}{12.2119} & \textcolor{blue}{11.3761} & \textcolor{blue}{10.9294} & \textcolor{blue}{48.5321} & \textcolor{blue}{42.5069} & \textcolor{blue}{40.2270} & \textcolor{blue}{38.6575} \\\cline{2-10}
    & $\mathcal{I}$ (\%) & \textbf{14.77} & \textbf{11.79} & \textbf{14.20} & \textbf{20.13} & \textbf{16.96} & \textbf{12.85} & \textbf{13.89} & \textbf{14.66}  \\ \hline
    
    \multirow{5}{*}{${\mathds{S}} + {\mathds{C}}  + {\mathds{O}}$}
    & DNM \cite{dnmm} & 18.4903 & 16.2861 & 15.1406 & 14.7933 & 63.2993 & 55.0821 & 49.3940 & 48.4284 \\
    & DCALF \cite{dcalf} & 17.6576 & 15.3595 & 14.3936 & 13.6697 & 51.4123 & 45.9013 & 42.6235 & 41.2194 \\
    & NCF \cite{ncf} & 15.4680 & 13.6160 & 12.2840 & 11.8330 & 49.7030 & 46.3040 & 42.3170 & 41.2630 \\
    & SPP+LLMF \cite{llmf} & 15.3820 & 13.6540 & 11.9040 & 	11.0890 & 51.5660 & 45.6810	& 41.6310 & 39.5340 \\
    & LDCF \cite{ldcf} & 13.8440 & 12.3820 & 11.2700 & 10.8410 & 47.3590 & 43.4820 & 39.8130 & 38.9980 \\
    & DCLG \cite{dclg} & \textcolor{blue}{12.9280} & \textcolor{blue}{11.3040} & \textcolor{blue}{10.4060} & \textcolor{blue}{9.9360 }& \textcolor{blue}{43.9690} & \textcolor{blue}{39.9600} & \textcolor{blue}{37.6370} & \textcolor{blue}{36.1760} \\\cline{2-10}
    & $\mathcal{I}$ (\%) & \textbf{-1.97} & \textbf{4.70}  & \textbf{6.20} & \textbf{12.15} & \textbf{8.34} & \textbf{7.30} & \textbf{7.97} & \textbf{8.80}  \\ \hline
    
   % &  UIPCC \cite{uipcc} & 25.8755 & 19.9754 & 17.5543 & 16.0762 & 60.8685 & 54.8761 & 47.8235 & 47.8749 \\\hline
    
    \multirow{1}{*}{${\mathds{S}} + {\mathds{GS}}  + {\mathds{O}} + {\mathds{C}}$}
    & \textbf{ARRQP} & \textbf{13.1839} & \textbf{10.7723} & \textbf{9.7604} & \textbf{8.7285} & \textbf{40.3005} & \textbf{37.0415} & \textbf{34.9693} & \textbf{32.9910}  \\\hline
    \multicolumn{10}{r}{* Improvement of ARRQP over the second best method highlighted in \textcolor{blue}{blue}}
 \end{tabular}
    \label{tab:tp_soa}
\end{table*}

\subsubsection{Comparison Metric} \noindent
To measure the performance of our framework, we used two performance metrics: Mean Absolute Error (MAE) and Root Mean Squared Error (RMSE) defined as follows:
\begin{equation}
    MAE = \frac{1}{|TD|} \sum_{(u_i, s_j) \in TD} |q_{ij} - {\hat{q}}_{ij}| 
\end{equation}
 and 
 \begin{equation}
     RMSE = \sqrt{\frac{1}{|TD|}\sum_{(u_i, s_j) \in TD}(q_{ij} - {\hat{q}}_{ij})^2}
 \end{equation}

 \noindent
 where $TD$ is the test dataset and $|TD|$ denotes the size of the test dataset. $q_{ij}$ and ${\hat{q}}_{ij}$ represent the actual and the predicted QoS values of a user-service pair $(u_i, s_j)$, respectively. 
 
In general, MAE is the average of non-negative differences between the actual and predicted sample that usually provides equal weight to all the samples. However, RMSE measures a quadratic score which gives relatively high weight to large errors. 

Furthermore, we used another measure, called \textit{Improvement}, to show the performance improvement of our framework over the past methods, which is defined below. 

\begin{definition}[Improvement $I (M_1, M_2) $] Given two error values $P_1$ and $P_2$ (measured in terms of MAE or RMSE) obtained by two different methods $M_1$ and $M_2$, respectively, the improvement of $M_1$ over $M_2$ is defined as:
\hfill$\blacksquare$
\begin{equation}
    {\mathcal{I}}(M_1, M_2)  = \left(({P_2 - P_1})/{P_2}\right) \times 100\%
\end{equation}
\end{definition}

\subsubsection{Parameter Configurations}
\label{subsec:config}
\noindent
To validate the performance of ARRQP, we used four different training-testing split-ups: $(5\%, 95\%), (10\%, 90\%), (15\%, 85\%), (20\%, 80\%)$. In this paper, we use the notation RT-$x$ (TP-$x$) to denote the training-testing split up for $\left(x\%, (100-x)\%\right)$. In other words, for RT-$x$, we used $x\%$ data of the given response time values for training, while the rest was used for testing. Moreover, we used 20\% of the training data as the validation dataset for our experiment. We repeated each experiment 10 times and reported the average value.

% We experiment split our dataset training densities, resulting in a total of eight distinct configurations. For further insight, Table \ref{tab:dataset_} provides a concise overview of the dataset details.

% \begin{table}[!h]\scriptsize
%     \centering
%     \caption{Training and test dataset details}
%     \begin{tabular}{c|c|c}
%      \hline%\hline
%      Parameter & Name  & Train : Test \\ \hline \hline
%      \multirow{4}{*}{RT} & RT-5 & $5\%:95\%$ \\
%      & RT-10  & $10\%:90\%$ \\
%      & RT-15  & $15\%:85\%$ \\
%      & RT-20  & $20\%:80\%$ \\ \hline
%      \multirow{4}{*}{TP} & TP-5 & $5\%:95\%$ \\
%      & TP-10  & $10\%:90\%$ \\
%      & TP-15  & $15\%:85\%$ \\
%      & TP-20  & $20\%:80\%$ \\ \hline
%     \end{tabular}
%     \label{tab:dataset_}
% \end{table}

In all our experiments, unless specifically stated otherwise, we maintained consistent usage of the parameters with values as listed in Table \ref{tab:paramters_values}.

\begin{table}[!h] \tiny
    \centering 
    \caption{Details of various parameters used in experiments}
    \begin{tabular}{c|c|c|c|c|c|c|c|c}  \hline
    Parameters & RT-5 & RT-10 & RT-15 & RT-20 & TP-5 & TP-10 & TP-15 & TP-20 \\ \hline \hline
    c, $\lambda$, t & \multicolumn{8}{c}{2, 0.1, 2} \\ \hline
    $d_n, d_s, d_c$  & \multicolumn{8}{c}{50, 50, 50} \\ \hline
    
    \multirow{2}{*}{$N_h$} \hspace{0.13cm} MAE & 5 & 1 & 4 & 3 & 5 & 4 & 5 & 5 \\ 
    \hspace{0.6cm} RMSE & 5 & 1 & 2 & 2 & 5 & 7 & 8 & 6  \\ \hline
    
    \multirow{2}{*}{$\gamma$} \hspace{0.35cm} MAE & 0.05 & 0.25 & 0.25 & 0.25 & 10 & 10 & 50 & 10 \\ 
    \hspace{0.6cm} RMSE & 250 & 10 & 100 & 10 & 10 & 50 & 50 & 100  \\ \hline
    \end{tabular}
    \label{tab:paramters_values}
\end{table}

\subsection{Experimental Results}\label{subsec:result_analysis}
\noindent
We now present the analysis of our experimental results.

\subsubsection{Comparison with State-of-the-Art Methods}
We compared ARRQP with 30 and 24 major State-of-the-Art (SoA) methods on RT and TP datasets, respectively. Tables \ref{tab:rt_soa} and \ref{tab:tp_soa} show the performance of ARRQP on RT and TP datasets for different training densities in terms of prediction accuracy, respectively. We have the following observations about the performance of ARRQP. 

\begin{itemize}[leftmargin=*]
    \item[(i)] {\emph{Comparison of prediction accuracy with SoA}}: As observed from Tables \ref{tab:rt_soa} and \ref{tab:tp_soa}, the prediction error started decreasing as the methods started handling more anomalies. It is worth noting that ARRQP outperformed the major SoA methods since it addressed more variations of anomalies compared to contemporary methods. The improvements of ARRQP over the other methods for each case (the methods are divided based on the anomalies addressed by them) are shown in the tables in bold. 

    \item[(ii)] {\emph{Comparison of prediction accuracy between ARRQP and DCLG}}: It may be noted from Table \ref{tab:tp_soa}, the performance of ARRQP degraded by $-1.97\%$ compared to DCLG \cite{dclg} in terms of MAE on TP-5. However, the performance of ARRQP improved by $8.34\%$ for the same in terms of RMSE. This shows that even though ARRQP increased the overall errors compared to DCLG, ARRQP was able to reduce the large prediction errors significantly. For the rest of the other cases, however, ARRQP outperformed DCLG in terms of MAE and RMSE.
    
    \item[(iii)] {\emph{Change of prediction accuracy with training density}}: As observed from Tables \ref{tab:rt_soa} and \ref{tab:tp_soa}, the performance of ARRQP improved with the increase in the training density. This is because, as the density increases, the number of neighbors of each node in the QoS prediction graph increases, which allows each user and service to have additional spatial collaborative information in the feature vector.
    
    \item[(iv)] {\emph{Training time of ARRQP}}: The training of ARRQP is performed in offline mode as discussed in Section \ref{sec:method}. The training time for ARRQP varied from 1.8 to 21.75 minutes, depending on the number of heads used in GCMF, which varied from 1 to 8 in our experiments. 
    
    \item[(v)] {\emph{Performance of ARRQP in terms of prediction time}}: The prediction time for ARRQP (including GRRQP and CRRQP) was in the order of $10^{-6}$ seconds, while the prediction time for SORRQP was in the order of $10^{-9}$ seconds. This difference in prediction time between SORRQP and ARRQP was because of the use of an additional MLP employed in GRRQP and CRRQP. The prediction time of ARRQP was negligible compared to the minimum service response time shown in Table \ref{tab:dataset_table}. The prediction of ARRQP is approximately 10 times faster than the prediction time for TRQP \cite{trqp}, which was in the order of $10^{-5}$ seconds, and it is the best-known among all the methods reported in this paper in terms of prediction time. 
\end{itemize}

\noindent
In summary, ARRQP achieved better prediction accuracy compared to the major SoA methods with a reasonable learning time and negligible prediction time, enabling it to integrate with a real-time service recommendation system.

We now discuss the performance of ARRQP addressing various anomalies discussed in Section \ref{sec:preliminaries}.

\subsubsection{Performance of ARRQP on Addressing Anomalies}
\noindent
This subsection analyzes various anomaly detection mechanisms used in this paper and the performance of ARRQP in dealing with those anomalies. We begin with discussing the performance of ARRQP handling outliers.

\begin{table*}[!h]
    \centering\scriptsize
    \caption{Performance of SORRQP after removal of outliers}
    \begin{tabular}{l|c|c|c|c|c|c|c|c|c}
        \hline
        \multirow{2}{*}{Methods} & Anomaly & \multicolumn{2}{c|}{RT-10} & \multicolumn{2}{c|}{RT-20} & \multicolumn{2}{c|}{TP-10} & \multicolumn{2}{c}{TP-20}  \\ \cline{3-10}
         & removed & MAE & RMSE & MAE & RMSE & MAE & RMSE & MAE & RMSE \\ \hline \hline
             LLMF\cite{llmf} & $10\%$ & 0.4041 & 0.6120 & 0.4037 &  0.6106 & 16.5509 & 33.8889 & 13.1105 & 27.9648 \\ 
             DALF\cite{DALF} & $10\%$ & 0.3955 & 0.7466 & 0.3439 & 0.6779 & 13.1968 & 27.8531 & 11.9619 & 26.0299 \\ 
            CAP\cite{cap} & $10\%$ & 0.3603 & 0.6439 & 0.3521 & 0.6640 & 16.4269 & 32.9558 & 16.3125 & 32.9334 \\ 
            TAP\cite{tap} & $10\%$ & 0.3385 & 0.5512 & 0.2843 & 0.4985 & 22.1419 &  43.4987 & 19.8273 & 40.9533 \\ 
            TRQP \cite{trqp} & $13\%$ & 0.2540 & - & 0.2520 & - & 10.5760 & - & 9.5660 & - \\ 
            OffDQ \cite{offdq} & $15\%$ & 0.2000 & - & 0.1800 & - & 9.1600 & - & 8.6700 & - \\ 
            CMF\cite{cmf} & $10\%$ & \textcolor{blue}{0.1762}& \textcolor{blue}{0.3705}& \textcolor{blue}{0.1524} & \textcolor{blue}{0.3599} & \textcolor{blue}{8.4573 }& \textcolor{blue}{24.9137} & \textcolor{blue}{7.2501} & \textcolor{blue}{20.8927} \\ 
%             MHwA & 10\% & 0.1202 & 0.2917 & 0.10705 & 0.26105 & 5.7032 & 14.57735 & 5.0371 & 14.29785\\ \hline
%             SHwoA & 10\% & 0.0963 & 0.3630 & 0.0744 & 0.2691 & 5.0767 & 18.5078 & 4.3573 & 17.4323  \\ \hline
\hline
            \textbf{SORRQP}  & \textbf{$10\%$} & \textbf{0.0930} & \textbf{0.3556} & \textbf{0.0794} & \textbf{0.2710} & \textbf{4.9652} & \textbf{18.3332} & \textbf{4.0876 }& \textbf{16.6755} \\ %\hline
\cline{1-2}            
            \multicolumn{2}{l|}{$\mathcal{I}$ (\%)} & \textbf{47.21} & \textbf{4.02} & \textbf{47.90} & \textbf{24.69} & \textbf{41.29} & \textbf{26.41} & \textbf{43.62} & \textbf{20.18} \\ \hline
            \multicolumn{10}{r}{* Improvement of SORRQP over the second best method highlighted in \textcolor{blue}{blue}}
    \end{tabular}
    \label{tab:anomaly_soa}
\end{table*}

\begin{figure}[!h]
 \centering
 \includegraphics[width=0.45\linewidth]{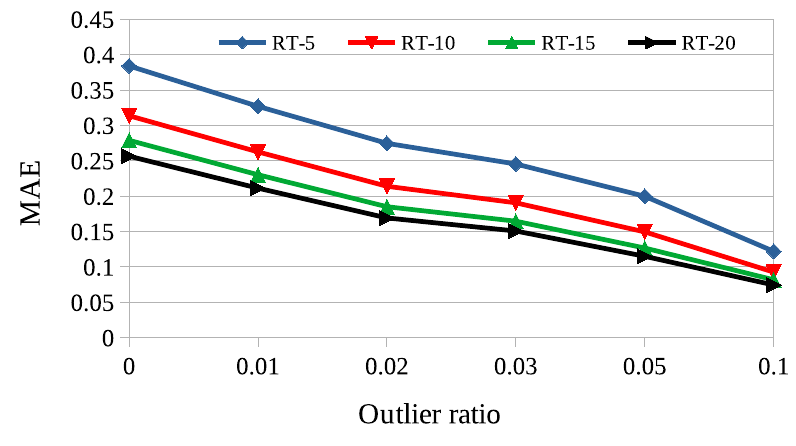}
 \includegraphics[width=0.45\linewidth]{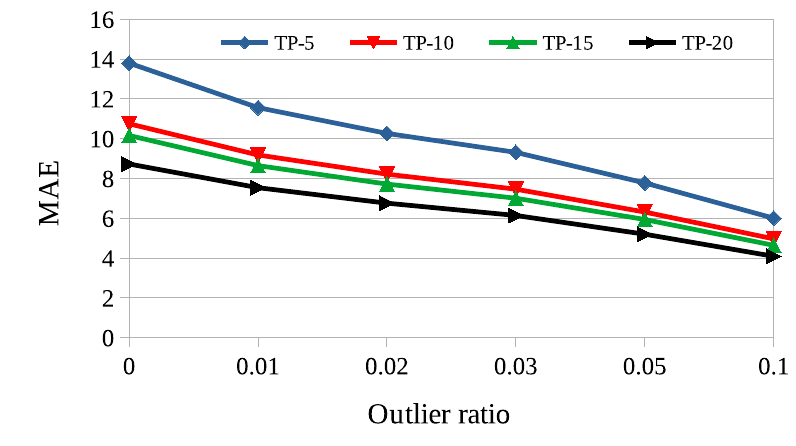}
 ~~~~~~~~~~~~~~~~~~~~~~~~~~~~~~~~~~(a)~~~~~~~~~~~~~~~~~~~~~~~~~~~~~~~~~~~~~~~~~~(b)~~~~~~~~~~~~~~~~~~~~~~~~~~~~~~~~~
 \caption{Analysis of outliers on (a) RT and (b) TP datasets}
 \label{fig:ol_detect}
\end{figure}

\noindent
\textbf{Impact of Outlier Detection Method: } Table \ref{tab:anomaly_soa} shows the performance of SORRQP after the removal of 10\% outliers (i.e., $\lambda = 0.1$). Here, we consider the SoA methods after removing $10\%$ or more data with anomalies, as mentioned in the second column of Table \ref{tab:anomaly_soa}. As observed from the final row of Table \ref{tab:anomaly_soa}, SORRQP outperformed all the major SoA methods in terms of prediction accuracy by a significant improvement margin.

Fig.s \ref{fig:ol_detect} (a) and (b) show the impact of the outlier detection algorithm used in this paper on RT and TP datasets, respectively.
Here, we showed the performance of our framework with the removal of $1\%, 2\%, 3\%, 5\% $ and $10\%$ outliers. As observed from Fig.s \ref{fig:ol_detect} (a) and (b), the performance of ARRQP (in terms of MAE) improved as more outliers were detected and removed, which shows the efficiency of the outlier detection algorithm.

\begin{figure}[!t]
    \centering
    \includegraphics[width=0.45\linewidth]{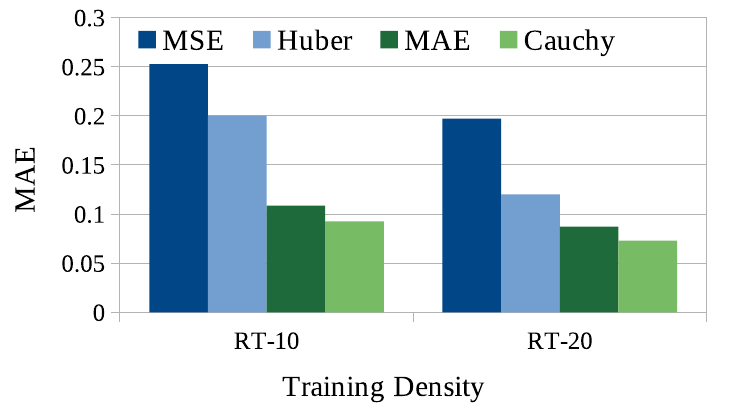}
    \includegraphics[width=0.45\linewidth]{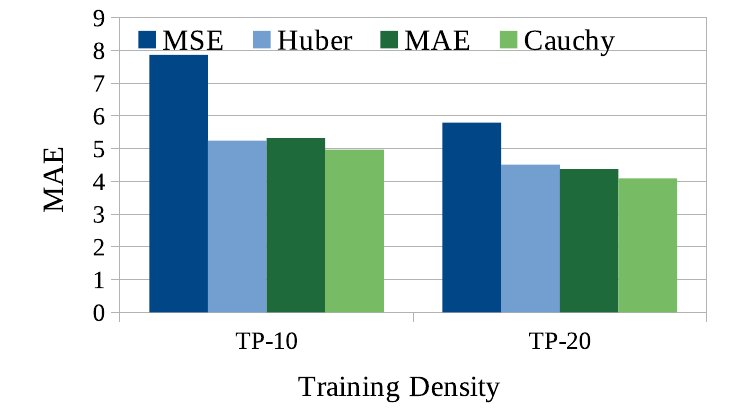}
    ~~~~~~~~~~~~~~~~~~~~~~~~~~~(a)~~~~~~~~~~~~~~~~~~~~~~~~~~~~~~~~~~~~~~~~~~(b)~~~~~~~~~~~~~~~~~~~~~~~~~~~~~~~~~
    \caption{Analysis of loss functions (a) RT and (b) TP datasets}
    \label{fig:loss}
\end{figure}

\noindent
\textbf{Impact of Loss Function to Handle Outliers: } 
Fig.s \ref{fig:loss}(a) and (b) show the impact of the loss functions on the performance of the ARRQP. The MSE is an outlier-sensitive loss function; therefore, it performed worst. MAE, Huber loss and Cauchy loss can deal with the outliers. In our framework, we adopt the Cauchy loss to train our model since it outperformed all other loss functions, as evident from Fig.s \ref{fig:loss}(a) and (b).

% % % % % % % % % % % % % % % % % % % % % % % % % % % % % % % % % % % % % % % % % 

\begin{figure}[!h]
 \centering
 \includegraphics[width=0.45\linewidth]{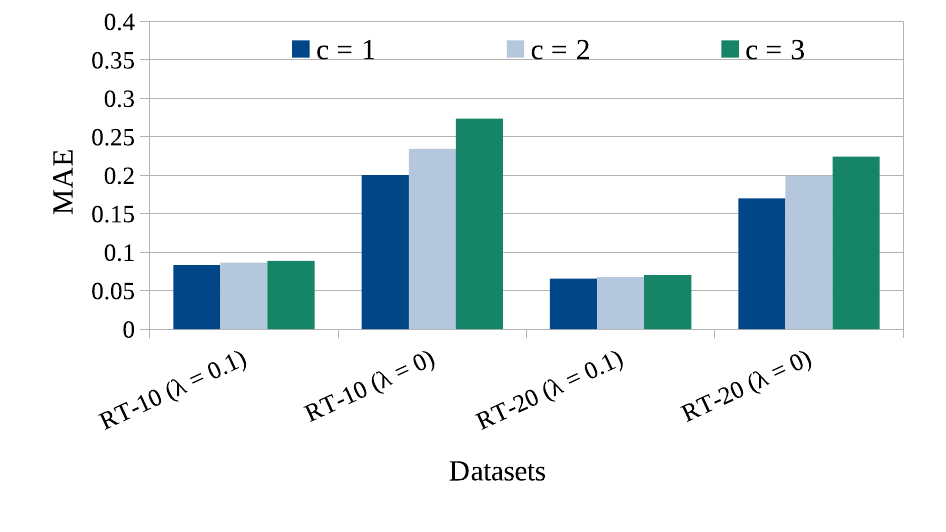}
 \includegraphics[width=0.45\linewidth]{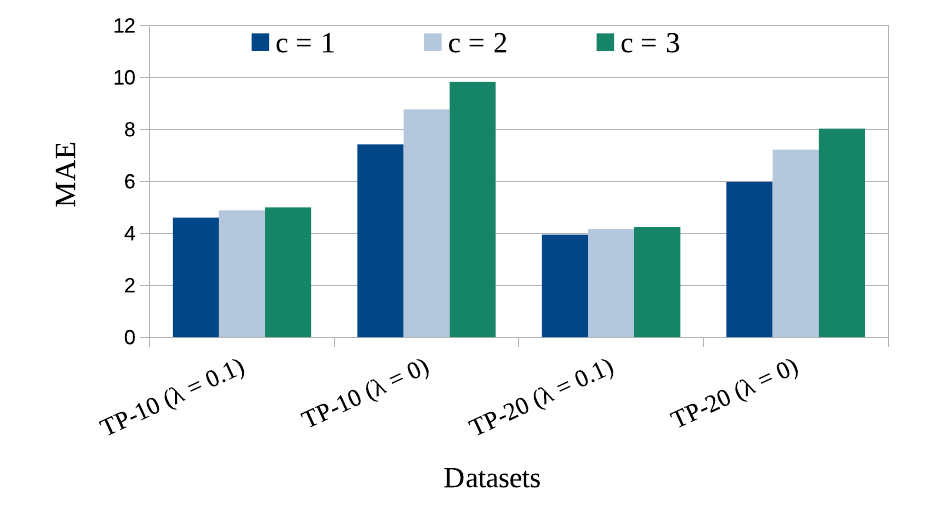}
 ~~~~~~~~~~~~~~~~~~~~~~~~~~~(a)~~~~~~~~~~~~~~~~~~~~~~~~~~~~~~~~~~~~~~~~~~(b)~~~~~~~~~~~~~~~~~~~~~~~~~~~~~~~~~
 \caption{Analysis of grey sheep detection metric (a) RT and (b) TP datasets}
 \label{fig:gs_detect}
\end{figure}

\noindent
\textbf{Impact of Grey sheep Detection Algorithm:} 
Fig.s \ref{fig:gs_detect} (a) and (b) show the Grey sheep detection capability of GRRQP on RT and TP datasets, respectively. We may infer the following observations from Fig.s \ref{fig:gs_detect} (a) and (b) below:
\begin{itemize}[leftmargin=*]
    \item [(i)] For $\lambda = 0$, as the value of $c$ decreased and the number of detected GSU and GSS increased and removed, the prediction error decreased. Table \ref{tab:gs_size} shows the number of detected GSU and GSS for different values of $c$.
    \item [(ii)] As we removed 10\% of the outliers along with the GSU and GSS for every value of $c$, the performance improvement of GRRQP was quite significant compared to the case with removing the GSU and GSS only without removing any outliers.
    \item [(iii)] For $\lambda = 0.1$, the decrease in the prediction error with a decrease in the value of $c$ was not that significant compared to the case for $\lambda = 0$. This finding suggests that the presence of a large number of outliers might contribute to the emergence of more grey sheep instances.
\end{itemize}

% Table \ref{tab:gs_size} show the datasets specifications for grey sheep analysis for $c = 2$, where $D_{GSU}$ and $D_{GSS}$ hold the GSU and all services and all users and grey sheep services, respectively.

\begin{table}[!b]
    \centering
    \scriptsize
    \caption{Number of (GSU, GSS) for different $c$}
    \begin{tabular}{c|c|c|c|c|c}
    \hline
        \multicolumn{2}{c|}{$c = 1$} & \multicolumn{2}{c|}{$c = 2$} & \multicolumn{2}{c}{$c = 3$} \\ \hline
         RT & TP & RT & TP & RT & TP \\ \hline
         (35, 254) & (24, 304) & (25, 150) & (7, 233) & (8, 101) & (4, 176)  \\ \hline
         % GSS & 254 & 304 & 150 & 233 & 101 & 176 \\ \hline 
        \end{tabular}
    \label{tab:gs_size}
\end{table}

\noindent
\textbf{Performance of GRRQP:} 
Table \ref{tab:gs_handling} presents the performance comparison between GRRQP and SORRQP to show the effectiveness of our framework in handling the GSU and GSS. While the SORRQP only addresses sparsity and outliers issues, the GRRQP offers explicit solutions for GSU and GSS in QoS prediction. This experiment was divided into two cases: (i) Case-1: considering all users and GSS $({\mathcal{U}}, {\mathcal{S}}_{{\mathcal{G}}})$, (ii) Case-2: considering GSU and all services $({\mathcal{U}}_{{\mathcal{G}}}{,\mathcal{S}})$. The improvement of GRRQP over SORRQP for Case-1 and Case-2 are shown in the last two columns of Table \ref{tab:gs_handling}.

% Notably, on average, GRRQP outperformed SORRQP by $1.25\%$ for GSU, $5.99\%$ for GSS, and $1.62\%$ for GSU, as well as $1.075\%$ for GSS, across both RT and TP datasets, respectively, as shown in Table \ref{tab:gs_handling}.

\begin{table}[!b]
 \scriptsize
 \caption{Performance (MAE) of GRRQP over SORRQP}
 \label{tab:gs_handling}
 \centering
 \begin{tabular}{c|c|c|c|c|c|c}
        \hline
        \multirow{2}{*}{Dataset} & \multicolumn{2}{c|}{SORRQP} &  \multicolumn{2}{c|}{GRRQP} & \multicolumn{2}{c}{$\mathcal{I}$ (\%)}\\\cline{2-7}
                & Case-1 & Case-2 & Case-1 & Case-2 & Case-1 & Case-2 \\ \hline\hline
        RT-10 & 1.7507 & 0.9278 & 1.7375 & 0.8857 & \textbf{0.75} & \textbf{4.54}\\ 
        RT-20 & 1.4828 & 0.6441 & 1.4568 & 0.5962 & \textbf{1.75} & \textbf{7.44}\\ 
        TP-10 & 109.3299 & 23.7551 & 107.7793 & 23.2875 & \textbf{1.42} & \textbf{1.97}\\ 
        TP-20 & 87.4062 & 20.215 & 85.8197 & 19.9767 & \textbf{1.82} & \textbf{1.18}\\ \hline
 \end{tabular}
\end{table}

\noindent
\textbf{Performance of ARRQP After Grey sheep Removal:} 
Table \ref{tab:gs_soa} demonstrates the performance of ARRQP after removing the data corresponding to GSU and GSS for $c = 2$.
Here, we compared our framework with the SoA methods that attempted to detect GSU and GSS and produced the results after removing them.
As observed from Table \ref{tab:gs_soa}, ARRQP outperformed all the major SoA methods reported in Table \ref{tab:gs_soa} with a significant improvement margin with or without removing outliers.

Moreover, it is worth mentioning that even though the SoA methods attempted to detect GSU and GSS, they failed to handle them. In contrast to the existing literature, ARRQP detects GSU and GSS and effectively handles them.

%%%%%%%%%%%%%%%%%%%%%%%%%%%%%%%%%%%%%%%%%%%%%%%%%%%%%%%%%%%%%%%%%%%%%%%%%%%%%%%%%%%%%%%%%%%%%%%%%%%%%%%%%%%%%%%%%%%

\begin{table*}[!h]    
    \begin{adjustbox}{width=1\textwidth}
    \parbox{.55\linewidth}{
   % {l|c|c|c|c|c|c|c|c}{p{1cm}|p{0.4cm}|p{0.4cm}|p{0.4cm}|p{0.4cm}|p{0.4cm}|p{0.4cm}|p{0.4cm}|p{0.4cm}}
    \centering\scriptsize
    \caption{{Comparison of ARRQP with SoA considering grey sheep removal}}
    \begin{tabular}{p{1cm}|p{0.6cm}|p{0.6cm}|p{0.6cm}|p{0.6cm}|p{0.6cm}|p{0.7cm}|p{0.6cm}|p{0.6cm}}
        \hline
        \multirow{2}{*}{Methods}  & \multicolumn{2}{c|}{RT-10} & \multicolumn{2}{c|}{RT-20} & \multicolumn{2}{c|}{TP-10} & \multicolumn{2}{c}{TP-20}  \\ \cline{2-9}
        & MAE & RMSE & MAE & RMSE & MAE & RMSE & MAE & RMSE \\ \hline \hline
        LRMF \cite{lrmf}	& \textcolor{blue}{0.4785} & \textcolor{blue}{1.2576} &  0.4340	& \textcolor{blue}{1.1422} & - & - & - & -  \\
        RAP	\cite{rap} &  0.4914 & 1.4346 & 0.4531	& 1.1631 & 21.4027 & - & 17.5978 & -  \\ 
        HLT	\cite{hlt} & 0.5088 & - & \textcolor{blue}{0.3634} & - & 28.5370 & - & 22.8296 & -  \\	
        CAP	\cite{cap} & 0.4997 & - & 0.4361 & - & \textcolor{blue}{16.6465} & -  & \textcolor{blue}{14.2685} & - \\
        TAP  \cite{tap} & 0.5978 & - & - & - & 25.7784 & - & - & - \\ 		\hline
        \textbf{ARRQP} & \multirow{2}{*}{\textbf{0.2343}} & \multirow{2}{*}{\textbf{1.0106}} & \multirow{2}{*}{\textbf{0.1992}} & \multirow{2}{*}{\textbf{0.9450}} & \multirow{2}{*}{\textbf{8.7754}} & \multirow{2}{*}{\textbf{32.1770}} & \multirow{2}{*}{\textbf{7.2216}} & \multirow{2}{*}{\textbf{28.7231}}\\ 
        {$\lambda = 0.0$} & &&&&&&&\\
        \hline
        {$\mathcal{I}$ (\%)} & \textbf{51.03}	& \textbf{19.64} &	\textbf{45.18}	& \textbf{17.26}	& \textbf{47.28} & \textbf{-} & \textbf{49.39} & \textbf{-} \\
        \hline
        \textbf{ARRQP} & \multirow{2}{*}{\textbf{0.2079}} & \multirow{2}{*}{\textbf{0.8180}} & \multirow{2}{*}{\textbf{0.1747}} & \multirow{2}{*}{\textbf{0.7475}} & \multirow{2}{*}{\textbf{8.1668}} & \multirow{2}{*}{\textbf{27.2536}} & \multirow{2}{*}{\textbf{6.7368}} & \multirow{2}{*}{\textbf{24.4762}} \\ 
        {$\lambda = 0.1$} & &&&&&&&\\
        \hline
        {$\mathcal{I}$ (\%)} & \textbf{56.55} & \textbf{34.95} & \textbf{51.93} & \textbf{34.56} &	\textbf{50.94}	& \textbf{-} &	\textbf{52.78} & \textbf{-} \\ \hline
    \multicolumn{9}{r}{* $\mathcal{I}$ of ARRQP ($c = 2$) over the second best method highlighted in \textcolor{blue}{blue}}
    \end{tabular}
    \label{tab:gs_soa}
    }
    \hfill
    \parbox{.4\linewidth}{
    \centering\scriptsize
    \caption{{Comparison of CRRQP with SoA considering the cold start for CSB on RT-10}}
    \begin{tabular}{p{1.3cm}|p{0.5cm}|p{0.5cm}|p{0.5cm}|p{0.5cm}|p{0.5cm}|p{0.5cm}}
        \hline
        \multirow{2}{*}{Methods}	& \multicolumn{3}{c|}{MAE} & \multicolumn{3}{c}{RMSE}\\\cline{2-7}
        & $0\%$ &	$5\%$ &	$10\%$ &	$0\%$ &	$5\%$ &	$10\%$ \\\hline\hline
        GMEAN &	1.0154 &	1.0300 &	1.0100 &	1.9678 &	1.9680 &	1.9702 \\
        USMF \cite{usmf} &	0.5499 &	0.5800 &	0.6100 &	1.4148 &	1.4954 &	1.5800 \\
        GeoMF1 \cite{geomf} &	0.4890 &	0.5200 &	0.5500 &	1.2022 &	1.3100 &	1.4000 \\
        GeoMF2 \cite{geomf} &	\textcolor{blue}{0.4610} &	\textcolor{blue}{0.5100} &	\textcolor{blue}{0.5200} &	\textcolor{blue}{1.1774} &	\textcolor{blue}{1.2900} &	\textcolor{blue}{1.3800} \\\hline
        
        \textbf{CRRQP} &	\textbf{0.3137} &	\textbf{0.3874} &	\textbf{0.4479} &	\textbf{1.1456} &	\textbf{1.2435} &	\textbf{1.2905} \\\cline{1-1}
        {$\mathcal{I}$ (\%)} & \textbf{46.95} & \textbf{31.64} & \textbf{16.09} & \textbf{2.77} & \textbf{3.73} & \textbf{6.93} \\\hline
        \multicolumn{7}{r}{GMEAN: uses the average QoS value to make predictions}
    \end{tabular}
    \label{tab:SoA_cs}
    }
    \end{adjustbox}
\end{table*}

\noindent
\textbf{Performance of CRRQP: } 
Cold start scenarios occur when new users or services are added to the system. Therefore, we divided our experiment into three cases as discussed below:

\begin{enumerate}[leftmargin=*]
    \item CSU: This is the first case where the experiment was designed with a set of cold start users without any data in the system. A cold start percentage (CSP) used for the experiment determines the number of cold start users. The experiment was limited to cold start users and all the services for this case.
    \item CSS: This is the second case. Here, the experiment was designed with a set of cold start services having no data in the system. Here CSP also determines the number of cold start services. The experiment was performed on all the users and cold start services for this case.
    \item CSB: This is the final case, which combines CSU and CSS. In this case, if the value of CSP is considered as $x$, the $x\%$ of the total users and $x\%$ of the total services are designated as the cold start users and services. The experiment here was performed jointly on all the users with cold start services and cold start users with all the services.
\end{enumerate}

\begin{figure}[!h]
 \centering
 \includegraphics[width=0.45\linewidth]{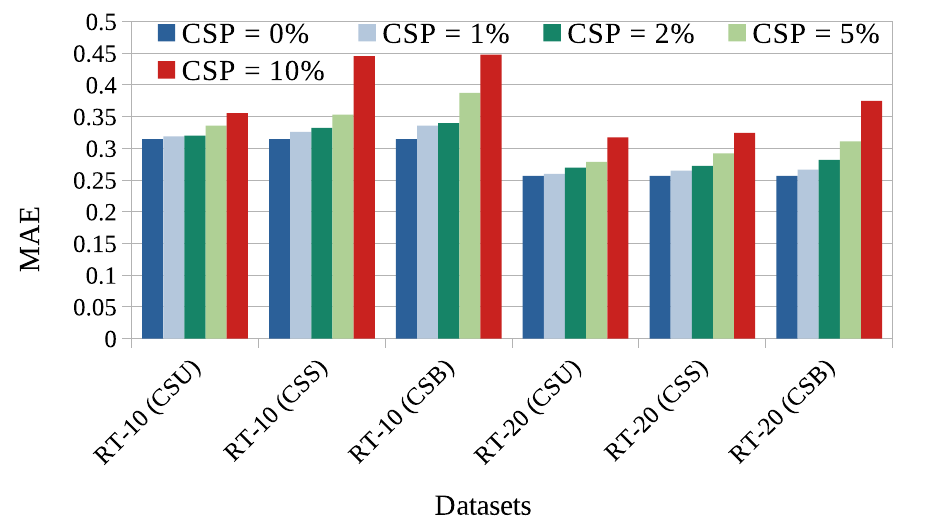}
 \includegraphics[width=0.45\linewidth]{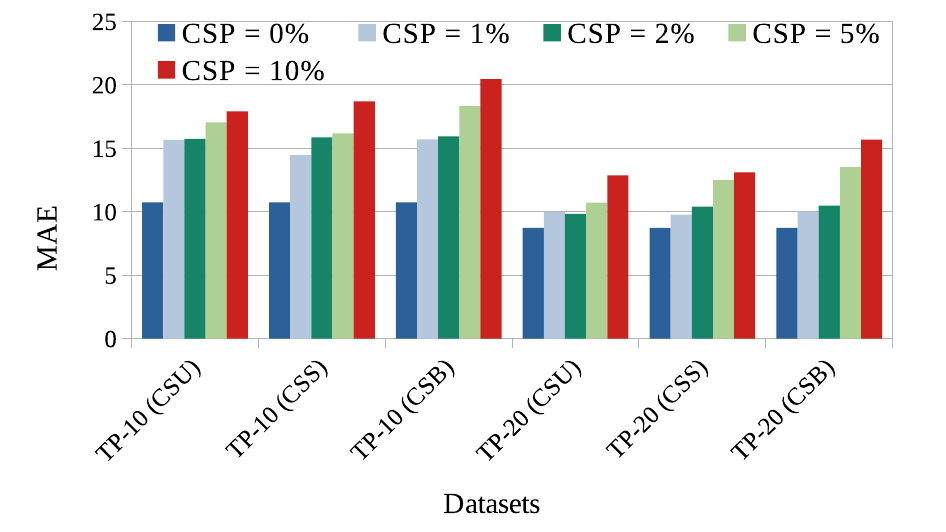}
 ~~~~~~~~~~~~~~~~~~~~~~~~~~~(a)~~~~~~~~~~~~~~~~~~~~~~~~~~~~~~~~~~~~~~~~~~(b)~~~~~~~~~~~~~~~~~~~~~~~~~~~~~~~~~
 \caption{Analysis of Cold-start on (a) RT and (b) TP datasets}
 \label{fig:cs_detect}
\end{figure}

\noindent
Here, we present the analysis for various values of CSP. 
Fig.s \ref{fig:cs_detect} (a) and (b) show the sensitivity of cold start on the performance of ARRQP on RT and TP datasets, respectively. As evident from Fig.s \ref{fig:cs_detect} (a) and (b), the performance of ARRQP deteriorated with the increasing value of CSP. Besides the unavailability of information on cold start users and services, the sparsity caused by the increasing CSP value was another reason for the performance degradation of ARRQP.
% \begin{itemize}
%     \item[(i)] 
%     \item[(ii)] For CSB case, the performance is worst compared to CSU and CSS cases. The reason behind this is that CSB represents an extreme cold start scenario, where the system has only contextual information available. In contrast, CSU has additional known services information, and CSS has additional known users information, making them comparatively less challenging than the CSB case.
% \end{itemize}

\begin{figure}[!b]
 \centering
\includegraphics[width=0.45\linewidth]{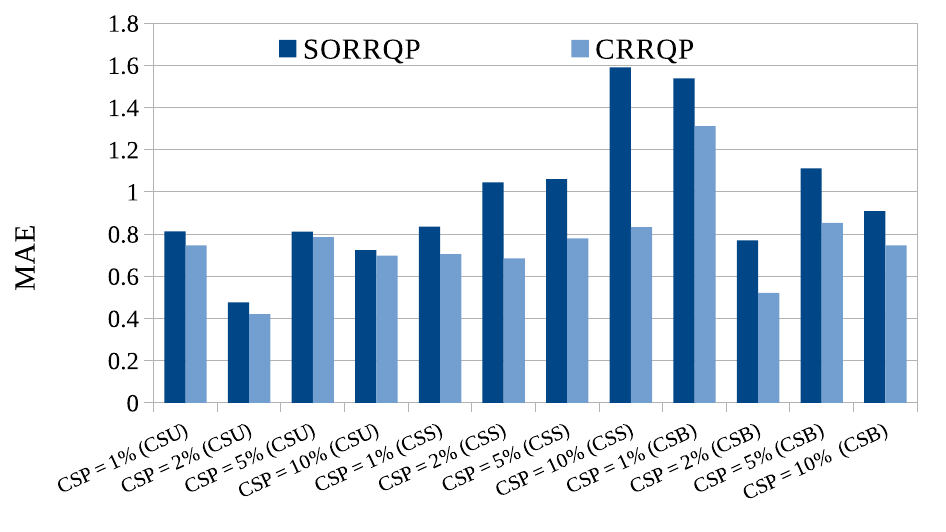}
 \includegraphics[width=0.45\linewidth]{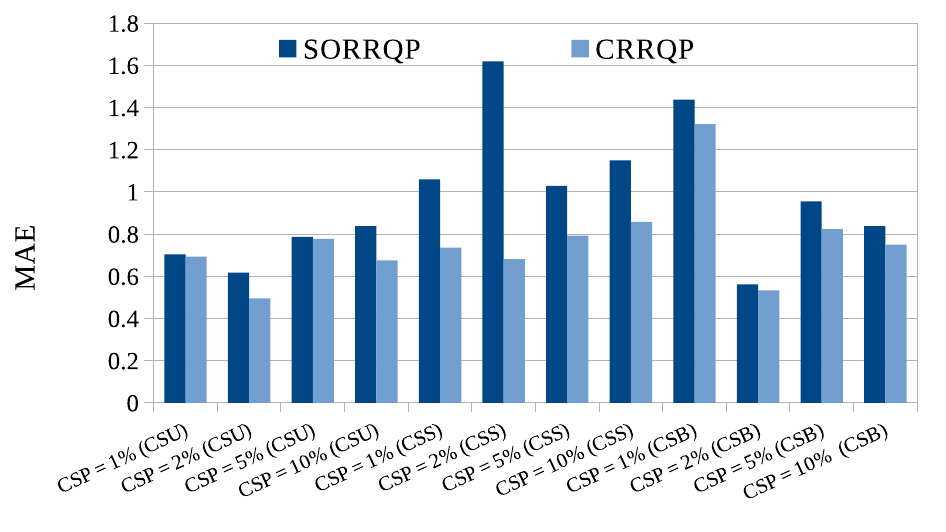}
~~~~~~~~~~~~~~~~~~~~~~~~~~~(a)~~~~~~~~~~~~~~~~~~~~~~~~~~~~~~~~~~~~~~~~~~(b)~~~~~~~~~~~~~~~~~~~~~~~~~~~~~~~~~\\
 \includegraphics[width=0.45\linewidth]{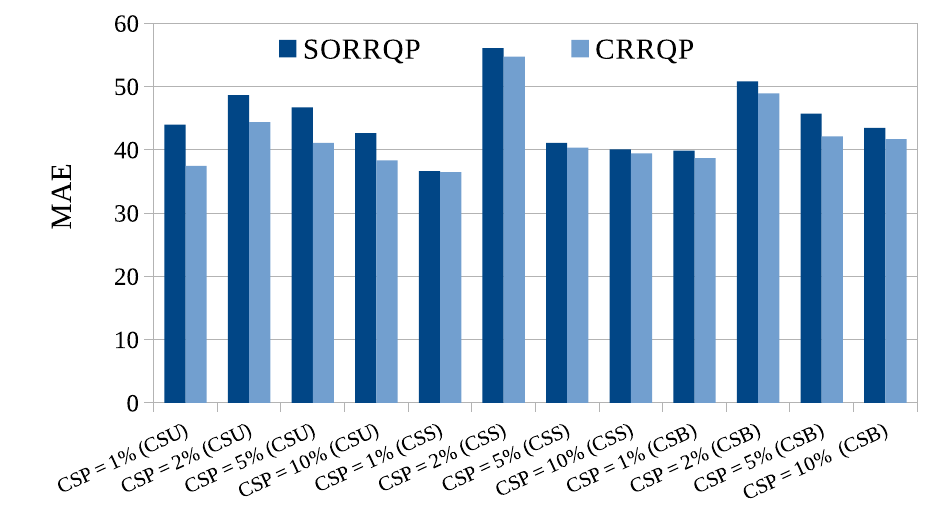}
 \includegraphics[width=0.45\linewidth]{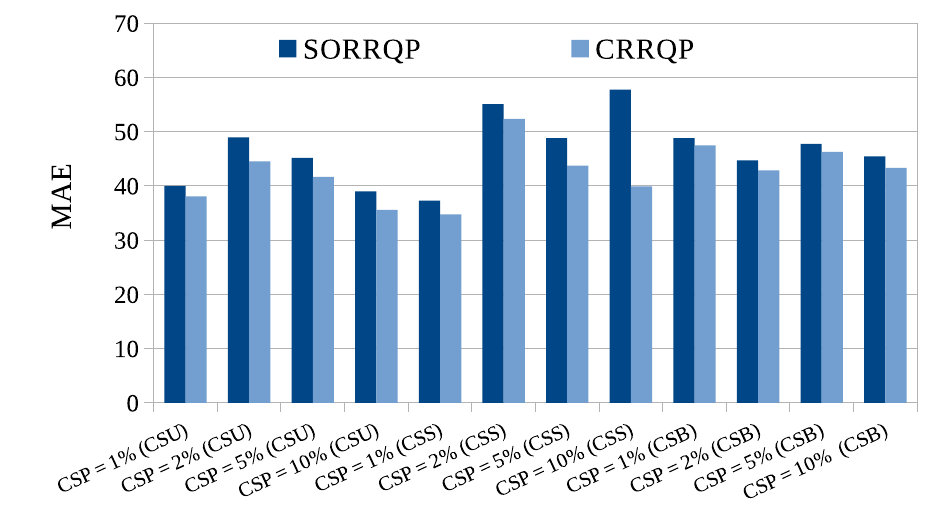}
 ~~~~~~~~~~~~~~~~~~~~~~~~~~~(c)~~~~~~~~~~~~~~~~~~~~~~~~~~~~~~~~~~~~~~~~~~(d)~~~~~~~~~~~~~~~~~~~~~~~~~~~~~~~~~
 \caption{Performance of CRRQP on (a) RT-10, (b) RT-20, (c) TP-10, (d) TP-20 datasets}
 \label{fig:cs_handling}
\end{figure}

Fig.s \ref{fig:cs_handling} (a)--(d) showcase the comparative performance of CRRQP and SORRQP on RT-10, RT-20, TP-10, and TP-20 datasets, respectively. As evident from Fig.s \ref{fig:cs_handling}(a)--(d), CRRQP consistently outperformed SORRQP, exhibiting a notable average improvement of $18.94\%$ and $7.04\%$ on RT and TP datasets, respectively. 

Table \ref{tab:SoA_cs} illustrates the performance of CRRQP compared to SoA methods for CSB with CSP of $0\%$, $5\%$, and $10\%$ on the RT-10 dataset. Table \ref{tab:SoA_cs} clearly demonstrates the significant performance enhancement achieved by CRRQP over the existing SoA methods that addressed the cold start problem.

The above studies show the efficiency of ARRQP in handling various anomalies. In the following subsection, we discuss the ablation study for our framework.
%%%%%%%%%%%%%%%%%%%%%%%%%%%%%%%%%%%%%%%%%%%%%%%%%%%%%%%%%%%%%%%%%%%%%%%%%%%%%%%%%%%%%%%%%%%%%%%%%%%
\subsubsection{Ablation Study}
We begin our discussion with the feature ablation study. 

\begin{figure}[!h]
 \centering
 \includegraphics[width=0.45\linewidth]{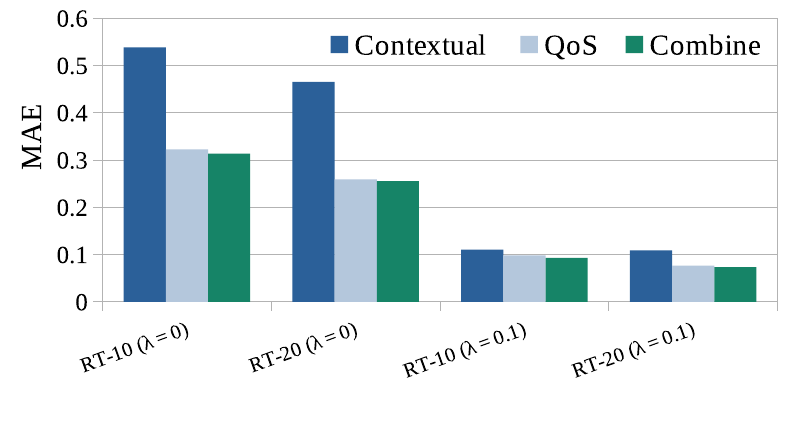}
 \includegraphics[width=0.45\linewidth]{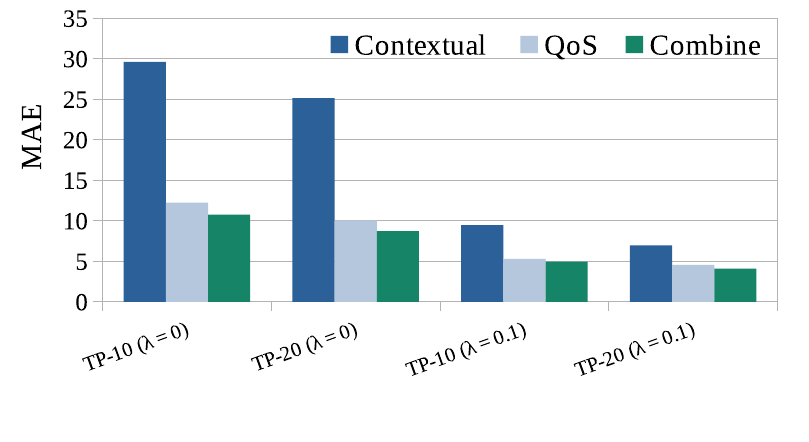}
~~~~~~~~~~~~~~~~~~~~~~~~~~~(a)~~~~~~~~~~~~~~~~~~~~~~~~~~~~~~~~~~~~~~~~~~(b)~~~~~~~~~~~~~~~~~~~~~~~~~~~~~~~~~\\ 
 \includegraphics[width=0.45\linewidth]{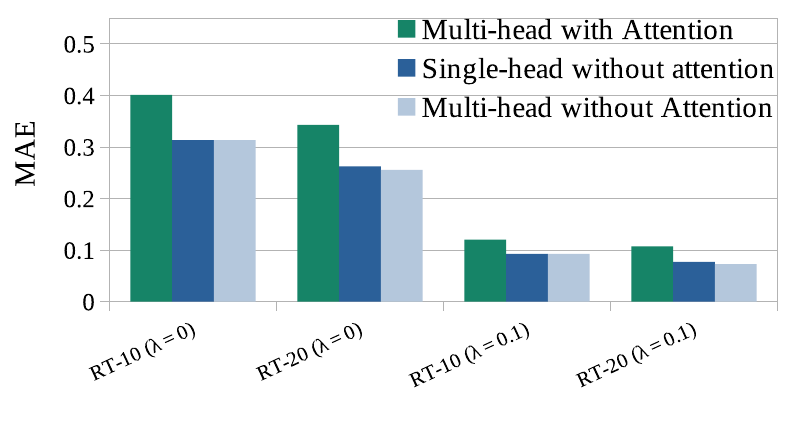}
 \includegraphics[width=0.45\linewidth]{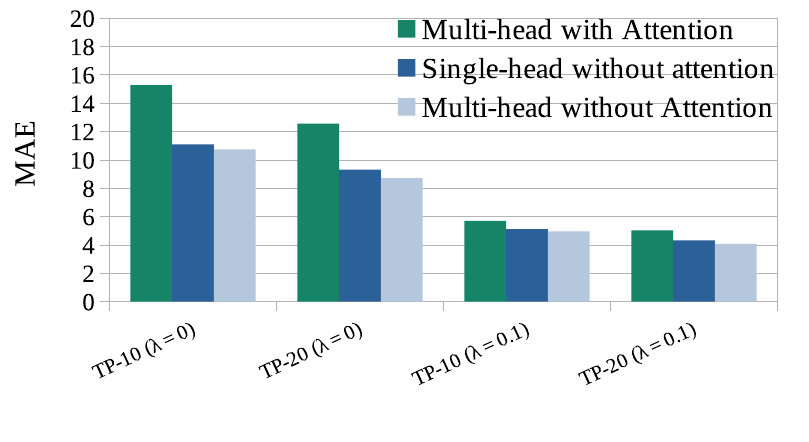}
 ~~~~~~~~~~~~~~~~~~~~~~~~~~~(c)~~~~~~~~~~~~~~~~~~~~~~~~~~~~~~~~~~~~~~~~~~(d)~~~~~~~~~~~~~~~~~~~~~~~~~~~~~~~~~
 \caption{Feature ablation study on (a) RT, (b) TP; Model ablation study on (c) RT, (d) TP datasets}
 \label{fig:ablation}
\end{figure}

\noindent
\textbf{Feature Ablation Study: }
Fig.s \ref{fig:ablation} (a) and (b) illustrate the influence of different features used in our framework. From Fig.s \ref{fig:ablation} (a) and (b), we can infer the following.
\begin{itemize}[leftmargin=*]
    \item[(i)] ARRQP with contextual features exhibited the worst performance. Even though the impact of QoS features was more prominent than the impact of the contextual features, however, ARRQP with a combination of contextual and QoS features outperformed the same with only QoS features by $2.01\%$ and $12.63\%$ on RT and TP datasets, respectively, on average.
    
    \item[(ii)] For $\lambda = 0.1$, all three models showed significant performance improvements compared to the case for $\lambda = 0$. The improvement percentages are  $78.07\%, 70.02\%, 70.93\%$ on RT, and $70.21\%, 55.75\%, 53.49\%$ on TP datasets for ARRQP with contextual features, QoS features and combined features, respectively.
\end{itemize}

\noindent
\textbf{Model Ablation Study: }
We studied the individual performance of GRRQP and CRRQP as compared to SORRQP in Table \ref{tab:gs_handling} and Fig.s \ref{fig:cs_handling} (a)--(d), respectively. These results emphasize the importance of GRRQP to handle grey sheep users and services and CRRQP to handle cold start situations in ARRQP alongside SORRQP to handle sparsity and outliers.

\subsubsection{Comparative Model Study} \label{subsec:model_study}
Here, we present a comparative study between three models: the single-head without attention model, the multi-head without attention model, and the multi-head with attention model. Fig.s \ref{fig:ablation} (c) and (d) present the performance of all three models on RT and TP datasets, respectively. The following can be inferred from Fig.s \ref{fig:ablation} (c) and (d):

\begin{itemize}[leftmargin=*]
    \item[(i)] The multi-head without attention model outperformed the multi-head with attention model \cite{gat} by an average improvement of $23.69\%$ and $30.08\%$ on RT and TP datasets, respectively. Moreover, it also surpassed the single-head without attention model by an average improvement of $1.34\%$ and $4.74\%$ on RT and TP datasets, respectively.
    \item[(ii)] A similar trend can be observed after the removal of 10$\%$ outliers.
\end{itemize}

%%%%%%%%%%%%%%%%%%%%%%%%%%%%%%%%%%%%%%%%%%%%%%%%%%%%%%%%%%%%%%%%%%%%%%%%%%%%%%%%%%%%%%%%%%%%%%%%%

\subsubsection{Impact of Hyper-parameters}
In this subsection, we study the influence of various hyper-parameters used in our framework.

\begin{figure}[!h]
 \centering
 \includegraphics[width=0.45\linewidth]{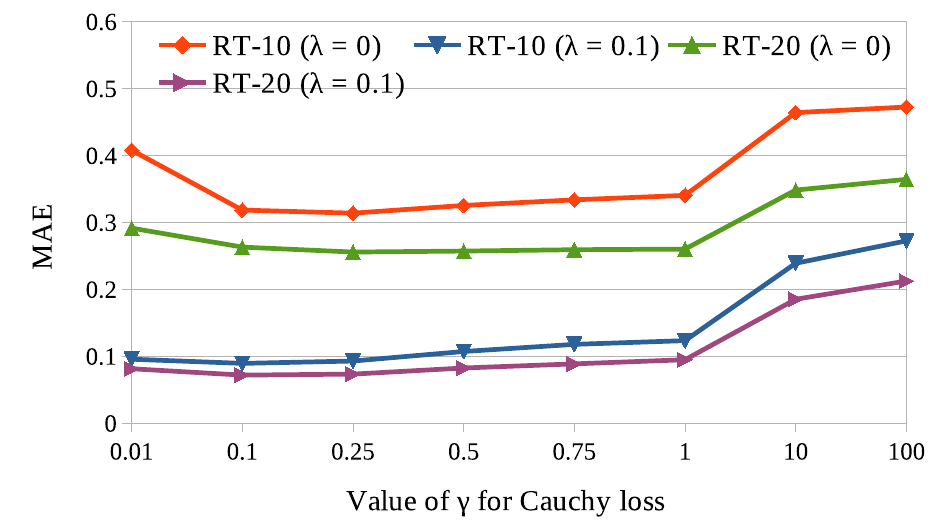}
 \includegraphics[width=0.45\linewidth]{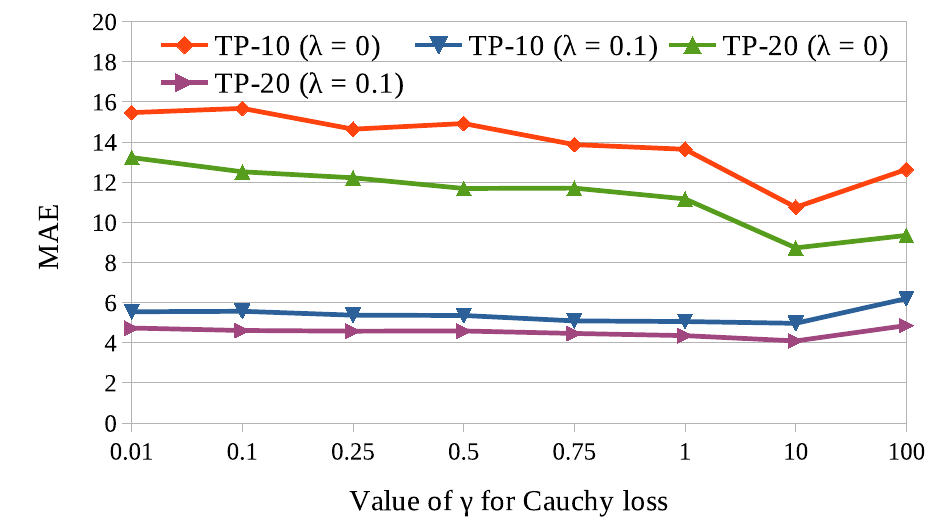}
 ~~~~~~~~~~~~~~~~~~~~~~~~~~~(a)~~~~~~~~~~~~~~~~~~~~~~~~~~~~~~~~~~~~~~~~~~(b)~~~~~~~~~~~~~~~~~~~~~~~~~~~~~~~~~\\
 \includegraphics[width=0.45\linewidth]{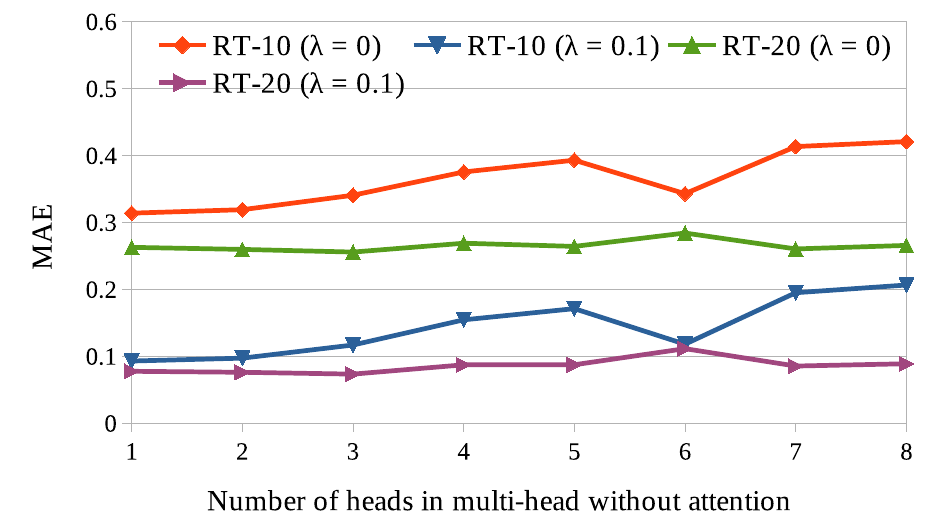}
 \includegraphics[width=0.45\linewidth]{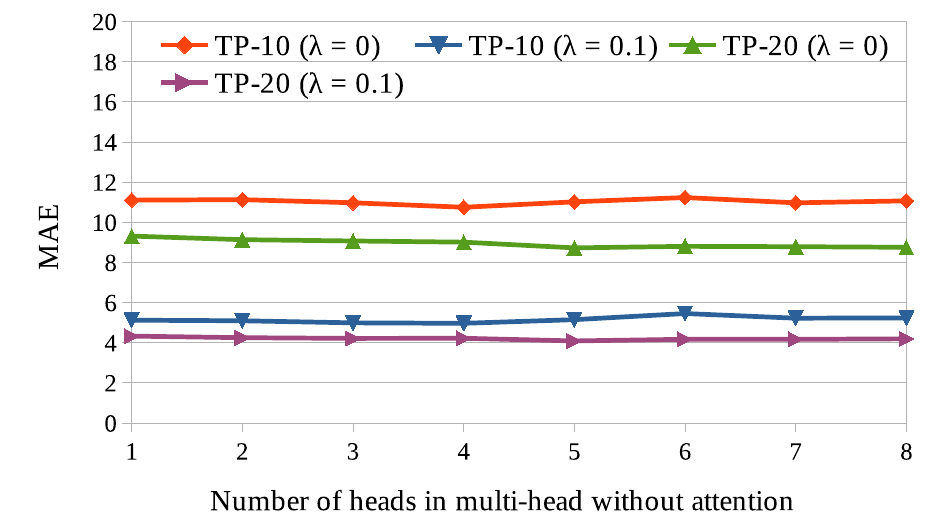}
 ~~~~~~~~~~~~~~~~~~~~~~~~~~~(c)~~~~~~~~~~~~~~~~~~~~~~~~~~~~~~~~~~~~~~~~~~(d)~~~~~~~~~~~~~~~~~~~~~~~~~~~~~~~~~
 \caption{Analysis of $\gamma$ on (a) RT and (b) TP; Analysis of number of heads in multi-head without attention model on (c) RT and (d) TP datasets}
 \label{fig:hyper_parameter}
\end{figure}

\noindent
\textbf{Impact of $\gamma$: }
$\gamma$ is a hyper-parameter present in the Cauchy loss function.
Fig.s \ref{fig:hyper_parameter} (a) and (b) illustrate the performance of ARRQP with the change in the value of $\gamma$ on RT and TP datasets, respectively. Among different $\gamma$ values, we observed that $\gamma = 0.25$ and $\gamma = 10$ achieved the best performance for all training densities on the RT and TP datasets, respectively. The same trend was followed after removing the $10\%$ of outliers from the datasets.

% \begin{figure}
%  \centering
%  (a) \includegraphics[width=0.4\linewidth]{fig/gamma_rt-eps-converted-to.pdf}
%  (b) \includegraphics[width=0.4\linewidth]{fig/gamma_tp-eps-converted-to.pdf}
%  \caption{Analysis of $\gamma$ on (a) RT and (b) TP datasets}
%  \label{fig:gamma}
% \end{figure}

\noindent
\textbf{Impact of Number of Heads: }
We tuned our model with different numbers of heads. The experimental results are shown in Fig.s \ref{fig:hyper_parameter} (c) and (d) for the RT and TP datasets, respectively. The following observations may be drawn from the figures: 
\begin{itemize}[leftmargin=*]
    \item[(i)] For RT datasets, although there is no fixed pattern, one head is sufficient to achieve the best performance at $10\%$ training density, while three heads performed best at $20\%$ training density. Moreover, removing $10\%$ of outliers enhances the model's performance.
    \item[(ii)] For TP datasets, the best performance is observed with four heads at 10\% training density and five heads at $20\%$ training density.
\end{itemize}

% \begin{figure}
%  \centering
%  (a) \includegraphics[width=0.4\linewidth]{fig/head_woa_rt-eps-converted-to.pdf}
%  (b) \includegraphics[width=0.4\linewidth]{fig/head_woa_tp-eps-converted-to.pdf}
%  \caption{Analysis of number of heads in multi-head without attention model on (a) RT and (b) TP datasets}
%  \label{fig:heads}
% \end{figure}

%%%%%%%%%%%%%%%%%%%%%%%%%%%%%%%%%%%%%%%%%%%%%%%%%%%%%%%%%%%%%%%%%%%%%%%%%%%%%%%%%%%%%%%%%%%%%%%%%%%%%%%%%

% To compute the CI, we utilized the entire test dataset, denoted as $|TD|$, and calculated the absolute differences from the ground truth. These resulting values were then randomly grouped into $g$ categories. Subsequently, we computed the Mean Absolute Error (MAE) for each group, and this information was employed in the CI calculation using Equation \ref{eq:ci}.

% \begin{equation}\label{eq:ci}
%     CI = \bar{x} \pm CL \frac{\sigma}{\sqrt{|TD|}}
% \end{equation}
% where $\bar{x}, \sigma$ and $|TD|$ denoted the sample mean, sample standard deviation and number of samples, which is the test dataset, respectively. 
\begin{table*}[!h] \scriptsize
    \centering
    \caption{Confidence Intervals}
    \begin{tabular}{c|c|c|c|c|c|c|c|c} \hline
        % \multirow{2}{*}{CL} & \multicolumn{4}{|c}{RT} & \multicolumn{4}{|c}{TP} \\ \cline{2-9}
         CL & RT-5 & RT-10 & RT-15 & RT-20 & TP-5 & TP-10 & TP-15 & TP-20 \\ \hline \hline
          90 & (0.3545, 0.3549) & (0.3052, 0.3055) & (0.2661, 0.2664) & (0.2399, 0.2402) & (12.5458, 12.5576) & (10.4042, 10.4152) & (8.8769, 8.8865) & (8.3365, 8.3450) \\
          95 & (0.3545, 0.3549) & (0.3051, 0.3055) & (0.2661, 0.2665) & (0.2398, 0.2402) & (12.5447, 12.5587) & (10.4032, 10.4162) & (8.8759, 8.8874) & (8.3357, 8.3458) \\
          99 & (0.3544, 0.3550) & (0.3051, 0.3056) & (0.2661, 0.2665) & (0.2398, 0.2402) & (12.5425, 12.5609) & (10.4011, 10.4183) & (8.8741, 8.8892) & (8.3341, 8.3474) \\ \hline
%           MAE & 0.3547 & 0.3053 & 0.2663 & 0.2400 & 12.5517 & 10.4097 & 8.8817 & 8.3408 \\
%           std\_dev & 0.0091 & 0.0076 & 0.0073 & 0.0070 & 0.2860 & 0.2672 & 0.2343 & 0.2076 \\ \hline
          % \multicolumn{6}{c}{MAE: Mean Absolute Error, STD: Standard Deviation}
    \end{tabular}
    \label{tab:ci}
\end{table*}

\subsubsection{Statistical Significance Testing}
To establish the reliability of ARRQP, we assess the statistical significance of our model using confidence intervals (CIs) \cite{pvalue}. A CI essentially provides a range around a measurement, indicating the precision of that measurement. Typically, CIs are computed for various confidence levels (CL), such as $90\%, 95\%$, and $99\%$. A $x\%$ CL implies that there is a $(100 - x)\%$ chance of uncertainty in the experiments for being wrong. 
% For our experiments, we opted for $g = 80$ as the grouping parameter to determine the CIs. Table \ref{tab:ci} showcases the CI experiments conducted for various CL on both the RT and TP datasets. In Table \ref{tab:ci}, the notation $x\%$ represents the training density on which the model was trained, while the CIs were computed on the test dataset (i.e., $100-x\%$). The rows corresponding to $90\%, 95\%$, and $99\%$ CL display the lower and upper bounds around the mean. Additionally, the table provides the mean and standard deviation in the last two rows.

The insights derived from Table \ref{tab:ci} enable us to evaluate the error precision of ARRQP under diverse conditions, facilitating well-informed decisions concerning the reliability of ARRQP. Notably, as we move from lower to higher CL, the CI tends to become wider.
In summary, this comprehensive analysis strengthens the validity of ARRQP and underscores its practical significance.

% To demonstrate the validity of their model, SoA methods typically employ one of two approaches: conducting extensive experimental studies involving a large number of samples or performing statistical significance tests, such as the Friedman test and Wilcoxon signed-ranks test (WSR) \cite{statistical_test}. However, it's essential to note that the p-value derived from statistical tests can be significantly influenced by large sample sizes, often leading to p-values approaching zero. Relying solely on p-values in such cases may not provide practical significance, as the low p-value could be a result of the sheer volume of data \cite{pvalue}. To address this limitation, it is advisable to complement statistical tests with experimental studies involving various sample sizes. This approach not only strengthens the credibility of the findings but also reveals practical significance. Notably, while DCALF \cite{dcalf}, GeoMF \cite{geomf}, PLF \cite{plf}, and D2E-LF \cite{d2elf} employ the WSR test, and SPP+LLMF \cite{llmf} use the T-test for pairwise significance testing with SoA methods, none of these methods report the effect size. In the literature \cite{pvalue}, it is recommended that when dealing with large samples, effect size should be reported using confidence intervals. This practice aligns with prior research helps mitigate the issue of low p-values driven primarily by large sample sizes.

\section{Related Work}\label{sec:related}
\noindent
This section presents a concise overview of relevant literature to discuss various anomalies in QoS prediction. In the domain of QoS prediction, Collaborative Filtering (CF) emerges as a widely adopted method.
CF-based methods predominantly rely on the collaborative relationships among users and services. They often leverage the similarity between users and/or services as a foundational element to facilitate QoS prediction. However, it is worth noting that CF-based methods often encounter various difficulties, including coping with high data sparsity, addressing cold-start issues, handling outliers, dealing with grey sheep problems, and exploiting intricate relationships among users and services. These challenges eventually result in elevated prediction errors \cite{zibinSurvey2022,ghafouriSurvey2022}. 
To address these challenges and enhance QoS prediction accuracy, more advanced techniques have been proposed in the literature. We now discuss various challenges and their possible solutions explored in the literature.
%%%%%%%%%%%%%%%%%%%%%%%%%%%%%%%%%%%%%%%%%%%%%%%%%%%%%%%%%%%%%%%%%%%%%%%%%%%%%%%%%%%%%%%%

\noindent\textbf{Data Sparsity:} 
Data sparsity is a prevalent challenge in the field of QoS prediction, which occurs due to an insufficient number of interactions between users and services, making it difficult to predict accurate QoS values because of the limited available information. To tackle the sparsity problem, the literature has proposed several solutions, including:
\emph{(i) Low-rank matrix decomposition:} It \cite{nmf,cmf,emf,pmf,llmf,snmf,rsnmf,csmf,nimf,ndmf,jcf} tackles the sparsity problem by capturing underlying patterns and relationships among users and services by decomposing the user-service interaction matrix into low-dimensional matrices and subsequently reconstructing it. However, while matrix decomposition-based methods are effective in handling sparsity, they encounter additional challenges, such as noise handling or difficulty in capturing higher-order relationships among users/services, potentially leading to inaccuracies in predictions.
% they may inadvertently capture noise instead of general patterns. Additionally, these methods assume linear relationships between users and services in the latent space, thereby, most of the time, they are unable to capture non-linear and complex features present in the data, resulting in low prediction performance. 
\emph{(ii) Designing additional data imputation method:} This is used sometimes to predict the missing values \cite{cadfm,offdq,cahphf,cnmf}. These imputed values are then utilized by the prediction module to estimate the final QoS value for the target user-service pair. However, the performance of the data imputation method directly affects the quality of predictions. Hence, the chosen data imputation method should be sophisticated to yield accurate results. Nonetheless, employing data imputation adds an extra layer of complexity to the prediction process, which, in turn, may result in reduced scalability and slower responsiveness of the prediction system.
\emph{(iii) Use of contextual data for prediction:} As seen in some approaches \cite{gmf,dnmm,lafil,lbr,msdae}, leveraging contextual data for prediction can help alleviate sparsity issues in certain cases. However, contextual data is not always readily available. Furthermore, when contextual information is used in isolation without any collaborative QoS data, it becomes challenging to capture the complex relationships between users and services. As a result, prediction accuracy may suffer and degrade. 
% The integration of both contextual and collaborative information enables a more comprehensive understanding of user-service interactions and leads to more accurate predictions.
%%%%%%%%%%%%%%%%%%%%%%%%%%%%%%%%%%%%%%%%%%%%%%%%%%%%%%%%%%%%%%%%%%%%%%%%%%%%%%%%%%%%%%%

\noindent\textbf{Presence of Outliers:} 
Outliers are data points that deviate significantly from the majority of the data, introducing anomalies that can hinder the performance of prediction algorithms. 
In the literature, outliers are predominantly addressed through two distinct approaches. {\emph{(i) Utilizing outlier-resilient loss functions:}} Some methods employ specialized loss functions that are resilient to the influence of outliers. These loss functions, such as L1 loss  \cite{dnmm,llmf,hsanet}, Huber loss \cite{hampel1986robust,dclg,ldcf}, Cauchy loss \cite{cmf,trqp}, have been demonstrated to be more effective than the standard L2 loss function when dealing with outliers. These robust loss functions downweight the impact of outliers during the training process, allowing the model to focus more on learning from the majority of the data.
{\emph{(i) Detecting and removing outliers:}} Alternatively, certain methods follow the explicit outliers detection \cite{offdq,cmf,trqp,dcalf} and followed by subsequent removal of them, resulting in a more reliable dataset for the prediction model.
%%%%%%%%%%%%%%%%%%%%%%%%%%%%%%%%%%%%%%%%%%%%%%%%%%%%%%%%%%%%%%%%%%%%%%%%%%%%%%%%%%%%%%%

\noindent\textbf{Presence of Grey sheep:} 
Grey sheep users/services refer to those with QoS invocation patterns that are sufficiently different from the others, often characterized by unique underlying behaviors. Therefore, the traditional and intuitive solutions may not effectively address them. The existing methods focus on identifying grey sheep instances and avoiding them. Here are some approaches used to detect the grey sheep.
RAP \cite{rap} mitigates the data credibility problem by leveraging the user reputation ranking algorithm, which identifies untrustworthy users present in the data. TAP \cite{tap} ensures credibility by employing unsupervised K-means clustering with beta distribution to calculate the reputation of users. CAP \cite{cap} utilizes a two-phase K-means clustering-based credibility-aware QoS prediction method, where clusters with a minimum number of users are considered untrustworthy. However, despite these efforts, data sparsity remains a significant challenge that can hinder the identification of grey sheep users or services. Sparse data may not provide enough information to accurately distinguish between typical and atypical behaviors, making it challenging to address grey sheep instances effectively.

Moreover, to the best of our knowledge, prediction methods tailored specifically for grey sheep users/services are mostly unexplored in the literature. However, understanding and effectively accommodating these grey sheep users/services in prediction algorithms can be crucial for achieving more accurate and personalized QoS predictions.
%%%%%%%%%%%%%%%%%%%%%%%%%%%%%%%%%%%%%%%%%%%%%%%%%%%%%%%%%%%%%%%%%%%%%%%%%%%%%%%%%%%%%%%%
\noindent\textbf{Cold Start:} 
The term {\emph{cold start}} is widely used to describe the scenario when new users or services are introduced to a system \cite{geomf}. These freshly added users or services typically lack data in the form of invocation history or have minimal associated data, leading to subpar prediction performance. It is important to note that the solutions designed to handle data sparsity are also highly relevant for addressing cold-start scenarios. Techniques developed to mitigate sparsity-related challenges can be adapted to improve predictions for cold-start instances by leveraging available data effectively \cite{lacf,lnlfm,namf,efm,ndmf,nimf,lafil,lbr,lemf,msdae,ldcf}.

Table \ref{tab:soa} presents a comprehensive summary of the literature on anomaly detection and handling, encompassing 38 SoA methods and our proposed framework. Notably, our framework demonstrates its efficacy in detecting and effectively handling all the mentioned anomalies.

\begin{table}[!h] \tiny
    \centering
    \caption{A brief literature survey on anomaly detection and handling of QoS prediction}
    \begin{tabular}{c|c|c|c|c|c|c}
    \hline
    \multirow{2}{*}{Methods} & \multicolumn{2}{c|}{Anomaly Detection} &   \multicolumn{4}{c}{Anomaly Handling} \\ \cline{2-7}
    & $\mathds{O}$ & $\mathds{GS}$ & $\mathds{S}$ & $\mathds{O}$ & $\mathds{GS}$ & $\mathds{C}$ \\\hline\hline
    
    \cite{upcc,ipcc,wsrec} &  & &  &  & & \\\hline

    % \cite{} &  & & \cmark &  & &  \cmark \\\hline

    \cite{nmf,pmf,snmf,lacf,lfm,lemf, nimf, lnlfm,rsnmf, lbr, usmf} &  & & \multirow{2}{*}{\cmark} &  & & \multirow{2}{*}{\cmark}\\
    \cite{namf,geomf,csmf,lmfpp,jcf,ndmf,efm,msdae,lafil,cnmf} &  & &  &  & & \\\hline

    % \cite{llmf} &  & & \cmark & \cmark & & \cmark\\\hline
    \cite{cap,rap,tap} &  & \cmark &  &  & & \\\hline
    \cite{offdq,cmf} & \cmark & & \cmark & \cmark &  & \cmark \\\hline
    % \cite{} & \cmark & & \cmark & \cmark & & \cmark \\\hline
    \cite{lrmf,hlt} &  & \cmark & \cmark &  & & \cmark \\\hline
    
    \cite{dcalf} & \cmark & \cmark &  & \cmark & &\\\hline 
    % \cite{,,} &  & & \cmark &  & &\cmark \\\hline
    
    \cite{llmf,dnmm,ldcf,dclg,hsanet} &   & & \cmark & \cmark & & \cmark \\\hline
    % \cite{} & & \cmark & \cmark &  & & \cmark \\\hline
    \cite{trqp} & \cmark & \cmark & \cmark & \cmark & & \cmark \\\hline
    \textbf{ARRQP} &  \cmark &\cmark &\cmark &\cmark &\cmark &\cmark \\\hline
    \end{tabular}
    
    \label{tab:soa}
\end{table}

\noindent
\textbf{Positioning Our Proposed Work:} 
In contrast to the approaches discussed above, this paper introduces several innovative solutions to mitigate the challenges associated with QoS prediction.

% \noindent
{\emph{(i)~}} We propose a simple yet effective solution leveraging graph convolution \cite{gcn}, which excels in dealing with high sparsity by aggregating and sharing neighborhood information of the nodes.

% \noindent
{\emph{(ii)~}}  Integrating both contextual information and QoS data, along with spatial collaborative information, offers a holistic perspective on user-service interactions here. This comprehensive approach enhances the system's ability to understand user behaviors and service performance, evidently resulting in more accurate predictions. 

% \noindent
{\emph{(iii)~}} The outlier-resilient Cauchy loss enables us to minimize the impact of outliers during the model training.

% \noindent
{\emph{(iv)~}}  The grey sheep detection block in ARRQP empirically shows the effectiveness in identifying grey sheep users/services. Importantly, it has the advantage of being able to handle the issue of data sparsity because it does not rely on measuring similarity between users/services. 

% \noindent
{\emph{(v)~}} In addition to the above, ARRQP offers an effective model specifically designed for predicting the QoS values of grey sheep users or services. 

% \noindent
{\emph{(vi)~}} The CRRQP block is specifically designed to address the prediction of QoS for newly added users/services. It recognizes the challenges posed by the cold start scenario, where these newcomers lack historical data, and takes special measures to make accurate predictions despite the limited information available. This specialized attention to cold start situations ensures that ARRQP can provide meaningful QoS predictions even for users or services with minimal or no prior usage history.

In summary, ARRQP demonstrates resilience to anomalies and is a well-suited framework for integration into real-time service recommendation systems, given its high scalability and responsiveness.

\section{Conclusion} \label{sec:conclusion}
This paper introduces an anomaly-resilient real-time QoS prediction framework (ARRQP), designed to achieve highly accurate QoS prediction in negligible time by addressing various challenges and anomalies, including the presence of outliers, data sparsity, grey sheep instances, and the cold start problem. ARRQP proposes a multi-head graph convolution matrix factorization model to capture complex relationships and dependencies among users and services. By doing so, it enhances its capacity to predict QoS accurately, even in the face of data limitations. Furthermore, ARRQP integrates contextual information and collaborative insights, enabling a holistic understanding of user-service interactions. Robust loss functions are employed to mitigate the impact of outliers during model training, improving predictive accuracy. In addition to QoS prediction, ARRQP introduces a sparsity-resilient method for detecting grey sheep users or services. These distinctive instances are subsequently handled separately for QoS prediction, ensuring tailored and accurate predictions.
Moreover, the cold start problem is addressed as a distinct challenge, emphasizing the importance of contextual features. ARRQP also exhibited a high responsiveness and scalability, rendering it well-suited for integration into real-time systems. This characteristic positions ARRQP as a valuable tool for applications where timely and efficient QoS prediction is essential.

The incorporation of advanced anomaly detection and mitigation algorithms holds promise for enhancing the performance of QoS prediction methods. Additionally, the development of a time-aware extension to ARRQP is a crucial area of focus for our future research endeavors. These directions represent our commitment to advancing the capabilities of QoS prediction and ensuring its relevance in dynamic and evolving systems.

\bibliographystyle{IEEEtran}
\bibliography{ref}

% \vspace{-0.5in}

\begin{IEEEbiography}[{\includegraphics[width=1in,height=1in,clip,keepaspectratio]{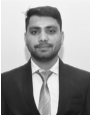}}]
{Suraj Kumar} is pursuing his PhD in CSE from the IIT Indore, India. He received his B.Tech and M.Tech in CSE from Aligarh Muslim University, India, in 2017 and 2019, respectively. His research interests include Services Computing, Machine Learning, and Graph Representation Learning.
\end{IEEEbiography}
\begin{IEEEbiography}[{\includegraphics[width=1in,height=1in,clip,keepaspectratio]{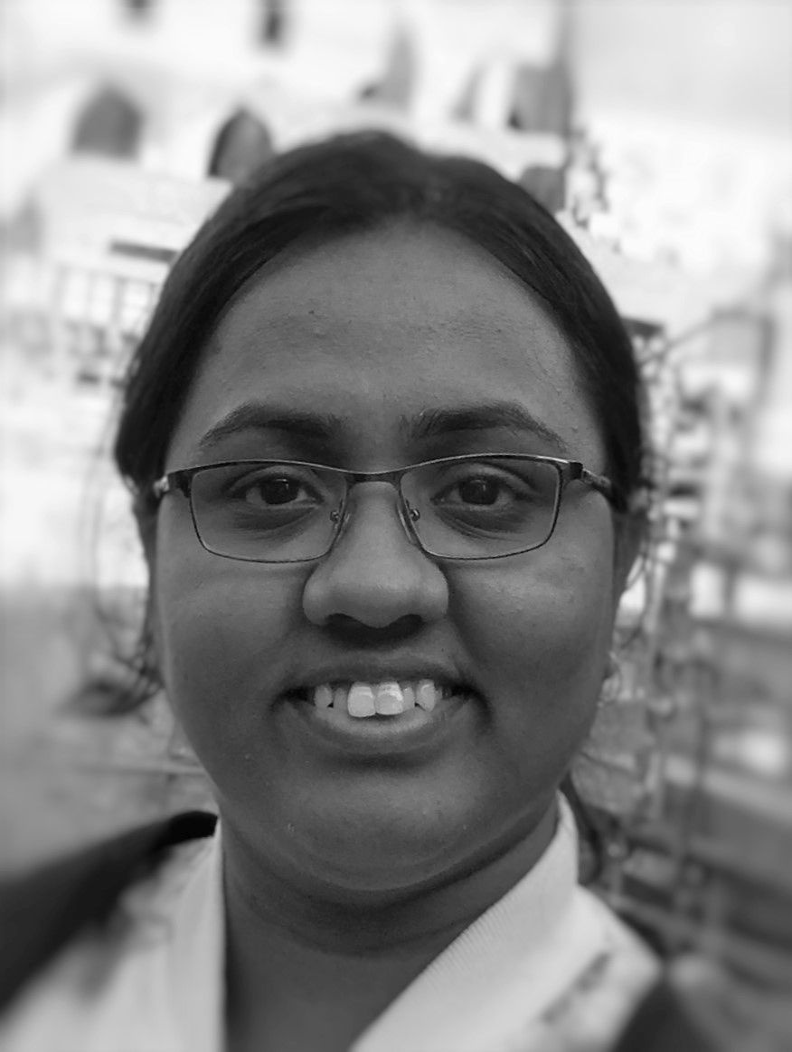}}]
{Soumi Chattopadhyay} (Member, IEEE) received her Ph.D. from the Indian Statistical Institute in 2019. Currently, she is an Assistant Professor at IIT Indore, India. Her research interests include Services Computing, Artificial Intelligence, Machine Learning, Deep Learning, Formal Languages, Logic and Reasoning.
\end{IEEEbiography}

% \newpage
% \input{7appendix}
\end{document}